\documentclass[10pt,journal,compsoc]{IEEEtran}
\usepackage{times}
\usepackage{epsfig}
\usepackage{graphicx}
\usepackage{amsmath}
\usepackage{amssymb}

\usepackage[utf8]{inputenc} 
\usepackage[T1]{fontenc}    
\usepackage{url}            
\usepackage{booktabs}       
\usepackage{amsfonts}       
\usepackage{nicefrac}       
\usepackage{microtype}      

\usepackage{multirow}
\usepackage{siunitx}
\usepackage{booktabs}
\usepackage{subfigure}
\usepackage[ruled,linesnumbered]{algorithm2e}

\SetCommentSty{mycommfont}
\makeatletter
\newcommand{\algorithmfootnote}[2][\footnotesize]{%
  \let\old@algocf@finish\@algocf@finish
  \def\@algocf@finish{\old@algocf@finish
    \leavevmode\rlap{\begin{minipage}{\linewidth}
    #1#2
    \end{minipage}}%
  }%
}
\makeatother
\usepackage[noend]{algpseudocode}
\graphicspath{ {fig/} }
\usepackage[shortlabels]{enumitem}
\usepackage{tabu}
\usepackage[flushleft]{threeparttable}

\newcommand{\type}[1]{\texttt{#1}}
\newcommand{\figref}[1]{Figure~\ref{#1}}
\newcommand{\secref}[1]{Section~\ref{#1}}
\newcommand{\tabref}[1]{Table~\ref{#1}}
\newcommand{\algoref}[1]{Algorithm~\ref{#1}}

\newcommand{\XX}[1]{{\textbf{#1}}}

\newcommand{\etal}{\emph{et al}.\xspace}

\DeclareMathOperator{\sgn}{sgn}

\usepackage[pagebackref=true,breaklinks=true,letterpaper=true,colorlinks,bookmarks=false]{hyperref}

\ifCLASSOPTIONcompsoc
  \usepackage[nocompress]{cite}
\else
  \usepackage{cite}
\fi
\ifCLASSINFOpdf
\else
\fi
\begin{document}
%
\title{NITI: Training Integer Neural Networks Using Integer-only Arithmetic}

%
%
%
%

\author{Maolin~Wang,
        Seyedramin~Rasoulinezhad,
        Philip~H.W.~Leong~\IEEEmembership{Senior~Member,~IEEE},
        and~Hayden~K.H.~So,~\IEEEmembership{Senior~Member,~IEEE}
\IEEEcompsocitemizethanks{\IEEEcompsocthanksitem Wang was with the Department
of Electrical and Electronic Engineering, University of Hong Kong, HKSAR. He is now with ACCESS – AI Chip Center for Emerging Smart Systems, InnoHK Centers, Hong Kong Science Park, Hong Kong, China. So is with the Department of Electrical and Electronic Engineering, University of Hong Kong, HKSAR. \protect\\
E-mail: \{mlwang,hso\}@eee.hku.hk
\IEEEcompsocthanksitem Rasoulinezhad and Leong are with the School of Electrical and Information Engineering, The University of Sydney.}
\thanks{Accepted to IEEE Transactions on Parallel and Distributed Systems (TPDS)
\href{http://dx.doi.org/10.1109/TPDS.2022.3149787}{[doi:10.1109/TPDS.2022.3149787]}}
}

%
%

\markboth{Wang \MakeLowercase{\textit{et al.}}: NITI: Integer Training}{}
%



\IEEEtitleabstractindextext{%
\begin{abstract}
  Low bitwidth integer arithmetic has been widely adopted in hardware implementations of deep neural network inference applications.
However, despite the promised energy-efficiency improvements demanding edge applications, the use of low bitwidth integer arithmetic for neural network training remains limited.
Unlike inference, training demands high dynamic range and numerical accuracy for high quality results, making the use of low-bitwidth integer arithmetic particularly challenging.
To address this challenge, we present a novel neural network training framework called NITI that exclusively utilizes low bitwidth integer arithmetic.
NITI stores all parameters and accumulates intermediate values as $8$-bit integers while using no more than $5$ bits for gradients.
To provide the necessary dynamic range during the training process, a per-layer block scaling exponentiation scheme is utilized.
By deeply integrating with the rounding procedures and integer entropy loss calculation, the proposed scaling scheme incurs only minimal overhead in terms of storage and additional computation.
Furthermore, a hardware-efficient pseudo-stochastic rounding scheme that eliminates the need for external random number generation is proposed to facilitate conversion from wider intermediate arithmetic results to lower precision for storage.
Since NITI operates only with standard $8$-bit integer arithmetic and storage, it is possible to accelerate it using existing low bitwidth operators originally developed for inference in commodity accelerators.
To demonstrate this, an open-source software implementation of end-to-end training, using native $8$-bit integer operations in modern GPUs is presented.
In addition, experiments have been conducted on an FPGA-based training accelerator to evaluate the hardware advantage of NITI.
When compared with an equivalent training setup implemented with floating point storage and arithmetic, NITI has no accuracy degradation on the MNIST and CIFAR10 datasets.
On ImageNet, NITI achieves similar accuracy as state-of-the-art integer training frameworks without relying on full-precision floating-point first and last layers.

\end{abstract}

\begin{IEEEkeywords}
 neural network training, integer arithmetic, NITI, GPU, Tensor Core, hardware accelerator
\end{IEEEkeywords}}

\maketitle

\IEEEdisplaynontitleabstractindextext

%
\IEEEpeerreviewmaketitle

\IEEEraisesectionheading{\section{Introduction}\label{sec:introduction}}
\IEEEPARstart{T}{raining} deep neural networks (DNNs) from scratch is a lengthy and computationally demanding process that requires a notoriously large number of arithmetic and memory operations. This forms a substantial barrier for rapid deployment and development of new applications and models.
Since the success of a DNN implementation depends heavily on the numerical accuracy of the training process, the use of $32$-bit single precision floating point arithmetic (\type{fp32}) has been the default standard for many DNN training frameworks and hardware systems.
However, as a continuous quest to improve performance, power and energy efficiency of training DNNs, both in edge scenarios and in cost-sensitive datacenters, there is significant interest in DNN training frameworks and accelerators that can operate with lower bitwidth or non-standard number systems~\cite{wage,wageubn}.
For instance, modern GPU-accelerated training frameworks already provide facilities to perform mixed-precision DNN training where half precision floating point numbers  (\type{fp16}) are used for most of the training process~\cite{mix-precision}.
In these cases, reducing the data bitwidth from $32$ to $16$ alone, allows larger models to be trained with better memory bandwidth utilization, and the smaller and more efficient hardware compute units enable improved power and more parallelism.

The challenge, however, is that the reduced dynamic range and accuracy of half precision floating point arithmetic can bring non-trivial degradation to the quality of DNN training. This effect is initialization and training set dependent.
As a result, alternative number representations, such as the brain floating point number (\type{bfloat16}) used in the tensor processing unit (TPU)~\cite{tpuv2}, and other mixed-precision schemes have been proposed~\cite{fp8-train-2018,fxpnet-2017}.
To reduce bitwidth to below \num{16}, Fox \etal recently proposed an $8$-bit mini block floating point scheme that they have demonstrated to successfully train a range of common neural networks with minimal accuracy loss~\cite{Fox:2021:ICLR}.  In a similar vein, using an $8$-bit Posit number system, Lu \etal have also demonstrated minimal degradation in training accuracy  compared to a \type{fp32} baseline~\cite{Lu:2021:TC}.  As a way to reduce energy consumption in edge applications, Lee \etal have also recently demonstrated a mixed \type{fp8}-\type{fp16} scheme in their neural network learning processor~\cite{Lee:2019}. 

While the above works have made promising progress toward low bitwidth DNN training, they all relied on non-standard floating point arithmetic that requires dedicated and sometimes non-existent hardware support.
As illustrated in \tabref{tab:datatypecmp}, the use of integer or fixed-point arithmetic promises to further reduce hardware complexity and improve energy efficiency when compared to their floating-point counterparts.
Furthermore, unlike non-standard floating-point operations, standard integer arithmetic units that operate on low bitwidth data are readily available in a wide range of existing computing systems and accelerators, making an integer-only DNN training framework an attractive alternative for rapid deployment especially in edge scenarios.
Unfortunately, training DNNs with integer-only arithmetic is particularly challenging because of their lack of dynamic range and precision that are needed throughout the training process.
Previous attempts in integer neural network training have commonly relied on floating-point arithmetic for at least a portion of the process to ensure successful training. Examples include accumulators, gradients, weights, error and activation computation during backpropagation~\cite{qnn-2017,8bit-integer-training,wageubn}.

In this work, we present NITI, a deep neural network training framework that operates \emph{exclusively} with integers, i.e. all parameters and intermediate accumulation values are updated and stored as $8$-bit integers.  To provide the necessary dynamic range for training, a per-layer block exponentiation scheme is proposed to dynamically adjust the scale of the stored values.  Furthermore, to address the unique challenges of training DNN with integers, 
a novel \emph{discrete} parameter update scheme that allows low precision integer storage of all intermediate variables is proposed.
To facilitate rounding of intermediate values to the low bitwidth storage, a hardware-efficient \emph{pseudo stochastic rounding scheme} that utilizes the extra precision in intermediate accumulation as an in-situ random number source has been developed.
Finally, an efficient approximation of cross-entropy loss backpropagation is proposed to allow integer-only arithmetic for even the challenging output layer of deep neural networks.

Since NITI operates exclusively with standard integer operations, it can readily be accelerated with existing accelerators that provide high performance integer operations.
To illustrate this, an open-source implementation of NITI accelerated with native \type{int8} matrix-multiplication using Tensor Cores of NVIDIA GPUs is presented.
To the best of our knowledge, the proposed implementation is the first to utilize the \type{int8} inference capability of modern GPUs to accelerate the entire training process without the help of any floating-point arithmetic.
Furthermore, to evaluate the hardware benefits of training neural networks using integer arithmetic only, 
an implementation of NITI's main algorithm using \type{int8} was compared to an equivalent  implementation that employed \type{fp32} operations on an FPGA.

In terms of training accuracy, we show that NITI can achieve negligible accuracy degradation on the MNIST dataset. 
On the CIFAR-10 dataset using a network with 9 weight layers similar to VGG~\cite{vgg}, also with \type{int8}, NITI was able to achieve \SI{91.8}{\percent} top-$1$ validation accuracy compared with \SI{91.5}{\percent} achieved by an equivalent floating-point implementation.  
On the ImageNet dataset using the original AlexNet, 
NITI was able to achieve \SI{48.3}{\percent} top-$1$ compared to \SI{49.0}{\percent} with an equivalent floating point implementation.

In the next section, we summarize related work in integer training.  The design of NITI will be presented in \secref{sec:overview}, followed by experimental results in 
\secref{sec:experiment}.  
The GPU and FPGA acceleration of NITI will be discussed in \secref{sec:acceleration}.
We will conclude and discuss future enhancements to NITI in \secref{sec:discuss}.

\section{Related Works}\label{sec:related}
\begin{table*}[htbp]
\caption{Datatype comparison for various low-precision integer DNN training frameworks}
  \label{table:related-low-precisoin}
  \centering
    \footnotesize
\begin{tabular}{|c|SSSSSc|}
    \hline
    & {$\boldsymbol{w}(infer)$}  &{$\boldsymbol{w}(acc)$} &{$\boldsymbol{a}$} & {$\boldsymbol{g}$} & {$\boldsymbol{e}$}  & {softmax} \\
    \hline
    TTQ\cite{ttq}  & 2 & 32   & 32 & 32 & 32 & {\type{fp32}} \\
    Xnor \cite{xnor-net}  & 1 & 32 & 1(32) & 32 & 32 & {\type{fp32}} \\
    Binaryconnect\cite{binary-net} & 1 & 32 & 32 & 32 & 32 & {\type{fp32}}   \\
    Jacob~\etal~\cite{gemmlowp-inference} & 8 & 32 & 8(32) & 32 & 32 & {\type{fp32}}  \\
    Dorefa\cite{dorefa} & 1 & 32 & 2(32) & 32 & 6(32) & {\type{fp32}}  \\
    Banner~\etal~\cite{8bit-integer-training} & 8 & 32 & 8(32) & 32 & 8(32) &{\type{fp32}}  \\
    WAGE\cite{wage} & 2 & 8(32) & 8(32) & 8(32) & 8(32) & {\type{fp32}} \\
    FxpNet\cite{fxpnet-2017} & 1 & 12(32) & 1(32) & 12(32) & 12(32) & {\type{fp32}}\\
    \cite{mix-int16} & 16 & 32 & 16 & 16 & 16 & {\type{fp32}} \\
    NITI (\textit{this work})  & 8 & 8 & 8 & 5 & 8 & integer \\
    \hline
    \multicolumn{7}{l}{(32) means the low precision data format is emulated by quantizing \type{fp32} number.}
\end{tabular}
\end{table*}

The study of hardware-efficient neural network training can be traced to research in low-precision neural network inference, which is a complementary problem.
Early studies of binary and ternary neural networks have already demonstrated the feasibility of using hardware-efficient bit-operations to replace complex floating-point multiply-accumulate (MAC) operations during training~\cite{binary-net,xnor-net,ttq}.
Moreover, researchers have also argued that such weight quantization during training could serve as a regularizer that improved the training performance on small datasets~\cite{binary-net}.
Subsequently, Jacob \etal extended quantization to $8$ bit weights and activations and demonstrated promising results for deploying DNN models on mobile devices~\cite{gemmlowp-inference}.
With a large-scale hardware accelerator, Gupta \etal~\cite{stochastic-rounding} further showed that DNNs could be trained with fixed-point weights using $16$-bit precision, and that rounding was crucial to success.
They proposed {\em stochastic rounding} where the probability of rounding $x$ to $\lfloor x \rfloor$ is proportional to their proximity, which has formed the basis of many modern integer training frameworks.

Beyond weight and activations, recent works have begun to
address the challenges of quantizing gradient and error computation during the backward pass of training.
In the work of DoReFa-Net~\cite{dorefa}, weight, activation as well as gradients of activations were quantized to allow discretized computation during backpropagation.
Similarly, Banners \etal developed a bifurcation scheme that quantized gradients of activations during training for $8$-bit integer operations in mobile processors~\cite{8bit-integer-training}.
Although promising results have been demonstrated in all the above cases, they have all relied on the use of full-precision floating-point computation in at least some parts of the training process.
In \cite{ttq,xnor-net,binary-net,gemmlowp-inference,dorefa,8bit-integer-training}, \type{fp32} were used as intermediate storage for weight accumulation, while in \cite{wage,fxpnet-2017} they were used as a proxy for the low-precision datatype for computation.

For efficient hardware implementations, a training scheme that employs integer arithmetic exclusively such as this proposed work is highly desirable. 
There are also platforms where a neural network training method without relying on floating-point is necessary. 
For example, \cite{memristor-cnn, memristor-mnist} explored training neural networks with memristor crossbar, whose underlying operations were integer arithmetic instead of floating-point arithmetic. 
Frequent switching between floating-point format and integer format would be an efficiency burden for memristor-based computation.   
Another example is federated machine learning at wireless edge \cite{ce-edge-learning, edge-ml}. In these works, gradients were calculated locally at edge devices that have strict power budgets. An integer-only training framework will save the circuit area reserved for the floating-point unit and leave more area for dedicated integer arithmetic engines, which can be used for both inference and training processes.

The closest works to NITI that we can identify are 
FxpNet~\cite{fxpnet-2017}, WAGE~\cite{wage} and ~\cite{mix-int16}. 
In ~\cite{mix-int16}, 16-bit integer arithmetic was used to accelerate forward pass, backward pass and weight update, 
but weight update still used \type{fp32} precision.
In FxpNet~\cite{fxpnet-2017}, $12$-bit fixed point arithmetic was employed throughout the training process to produce a binary network for inference. 
However, their proposed framework was only tested on CIFAR10 and lacked the ImageNet results.
In WAGE, weight, activations, gradients, and error values were all quantized to $8$-bit integers to train a ternary network for inference.  
They achieved moderate accuracy drop using 8 bits to train AlexNet on ImageNet, but the first convolution layer, last fully connected layer, and softmax still uses \type{fp32} for computation. 
In contrast, NITI employs integer arithmetic exclusively, even when used in training large networks for classifying ImageNet dataset with \num{1000} classes using softmax and have achieved only moderate accuracy degradation.  
A summary of the datatype employed in related works is shown in \tabref{table:related-low-precisoin}.

The goal of NITI is to facilitate efficient hardware accelerator designs for neural network training,  particularly in demanding scenarios where area and power-efficiency (throughput/watt) are the dominating optimization goals such as in the edge and in remote deployment locations. 
From \tabref{tab:datatypecmp} below, 
it is evident that all floating point datatypes (\type{fp8}, \type{fp16}, \type{bfloat16}) consume significantly more resources than \type{int8}. implementations: \num{3.1}$\times$ to \num{7.2}$\times$ more LUTs and 
\num{4.3}$\times$ to \num{7.4}$\times$ more registers in FPGAs; 
\num{2.2}$\times$ to \num{4.3}$\times$ more silicon area in ASICs, which is linearly proportional to power-efficiency. ($\text{power-efficiency}=\text{power}/\text{freq}=k\times \text{area}$,  where $k$ is a constant.) 
Even when the additional storage needed for scaling factors and extended intermediate parameters are taken into account, the significant advantages in area and power-efficiency still present a strong case for \type{int8}-only training accelerators.
Moreover, as illustrated by our current implementation using the Tensor Cores in nvidia GPUs, \type{int8} operations are readily available in today’s computer systems, both in the high performance and low power areas, often accelerated by SIMD/vector extensions and GPUs.  NITI enables accelerated training on these standard platforms by using \type{int8} exclusively.
\begin{table*}[htb]
    \centering
    \footnotesize
    \caption{Preliminary results of multiply-accumulate (MAC) operator hardware resource consumption using low-precision datatype proposed in related works.  Designs generated with deepfloat framework\cite{johnson2018rethinking}, synthesized for Xilinx FPGAs using Vivado, and synthesized for ASIC with Yosys targeting FreePDK45nm process.  Floating point designs pipelined to maintain reasonable clock speed.  
    Generic \type{fp16} and \type{bfloat16} configuration were used for accumulation.  Floating point data configuration $(\text{s}, \text{e}, \text{m})$ stands for the number of bits devoted to sign, exponent and mantissa.
}
    \label{tab:datatypecmp}
    \begin{tabular}{cccccccc}
        \toprule
Datatype&Input Data&Accumulate Data&\multicolumn{3}{c}{FPGA Implementations}&\multicolumn{2}{c}{ASIC Implementations}\\
&Config&Config&LUT&Reg&freq (\si{\mega\hertz})&area (\si{\micro\meter\squared})&freq (\si{\mega\hertz})
\\
        \midrule
\type{fp16}&(1,5,10)&(1,5,10)&506&152&528&1412&1077
\\
\type{bfloat16}&(1,8,7)&(1,8,7)&419&152&449&1165&1766
\\
\type{fp8} \cite{fp8-train-2018}&(1,5,2)&(1,5,10)&362&136&556&980&1522
\\
\type{fp8}\cite{NEURIPS2019_65fc9fb4} &(1,4,3)&(1,6,9)&395&140&554&1065&1579
\\
\type{fp8} \cite{Cambier2020} &(1,5,2)&(1,8,23)&886&238&425&1972&1096
\\
\type{int16}&\type{int16}&int32&345&32&775&903&1001
\\
\type{int8} (NITI)&\type{int8}&int32&120&32&775&455&1001
\\
        \bottomrule
    \end{tabular}
    
\end{table*}
\section{An Integer-only Training Framework}\label{sec:overview}

\begin{algorithm}[tb]
\label{alg:int-train}
\caption{Forward and backward passes in NITI for each batch of data $X$ and label $Y$}
     \tcc{Forward Pass, with quantized input $\boldsymbol{a}^{(0)}, s_{a^{(0)}}$ from $X$}
     \For{\textup{each layer $l$}}{
     $ \boldsymbol{a}_{\texttt{32}}^{(l)},\: s_{a^{(l)}}
     \gets$
     \Call{Int8MatrixMultiply}{$\boldsymbol{a}^{(l-1)},\boldsymbol{w}$}$,\: s_{a^{(l-1)}}+s_w$\;
     $b \gets$\Call{EffectiveBitwidth}{$\boldsymbol{a}_{\texttt{32}}^{(l)}$}\;
     $\boldsymbol{a}^{(l)}, s_{a^{(l)}} \gets $ \Call{ShiftAndRound}{$\boldsymbol{a}_{\texttt{32}}^{(l)}, s_{a^{(l)}}, \max(0, b-7)$}\nllabel{algln:sara}\;
    }
    \tcc{Backward Pass, with $\boldsymbol{a},s_a$ being final layer's output, scale}
    $\boldsymbol{e} \gets$
    \Call{Int8LossGradient}{$\boldsymbol{a},s_a,Y$}\;
    \For{\textup{each layer $l$ in reverse order}}{
    $ \boldsymbol{g}^{(l)}_{\texttt{32}} \gets \Call{Int8MatrixMultiply}{\boldsymbol{a}^{(l-1)},\boldsymbol{e}^{(l)}}$\;
    $ \boldsymbol{e}^{(l-1)}_{\texttt{32}} \gets \Call{Int8MatrixMultiply}{\boldsymbol{w}^{(l)},\boldsymbol{e}^{(l)}}$\;
    $b \gets$\Call{EffectiveBitwidth}{$\boldsymbol{e}^{(l-1)}$}\;
    $\boldsymbol{e}^{(l-1)}, \_ \gets $ \Call{ShiftAndRound}{$\boldsymbol{e}_{\texttt{32}}^{(l-1)}, \_, \max(0, b-7)$}\nllabel{algln:sare}\;

    $b \gets$ \Call{EffectiveBitwidth}{$\boldsymbol{g}^{(l)}_{\texttt{32}}$}\;
    $\boldsymbol{g}^{(l)},\_ \gets$ \Call{ShiftAndRound}{$\boldsymbol{g}^{(l)}_{\texttt{32}},\_,\max(0, b-m_u)$}\;
    $\boldsymbol{w}^{(l)} \gets \boldsymbol{w}^{(l)}-\boldsymbol{g}^{(l)}$ \;    
    }
\end{algorithm}

\algoref{alg:int-train} shows the overall training algorithm of NITI, which is based on the commonly used stochastic gradient descent (SGD) with backpropagation (BP) scheme.
To achieve the goal of computing and storing all values as integers, NITI relies on $3$ tightly coupled innovations: (i) a per-layer block exponentiation scaling scheme for integer values; (ii) a discrete weight update scheme; and (iii) a shift-and-round scheme for intermediate values that integrates with the above two.

\begin{figure*}[tbp]
    \centering
    \includegraphics[width=\linewidth]{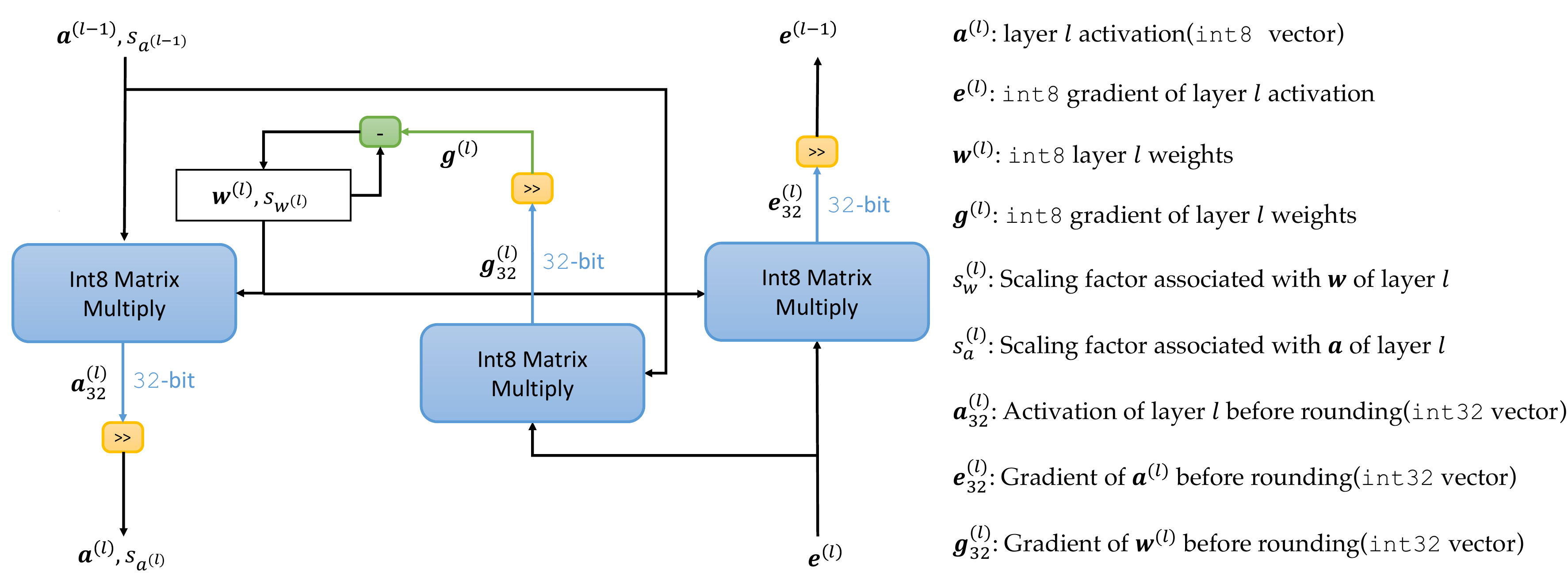}
    \caption{Integer layer forward and backward pass}
    \label{fig:intflow}
\end{figure*}

The judicious use of datatype is essential to the success of NITI.   \figref{fig:intflow} shows the mathematical symbols and their associated datatypes for various parts of the NITI algorithm.  To facilitate discussions, throughout this work, vectors are named with lower case alphabets and are typeset in boldface (e.g. $\boldsymbol{a}$) while matrices are named with upper case alphabets (e.g. $\boldsymbol{X}$).
Scalars are typeset with normal fonts (e.g. $a$) with an optional parentheses in superscript (e.g. $a^{(l)}$) denoting the layer $l$ that it belongs.  Furthermore, unless otherwise noted, all scalar variables are represented as $8$-bit signed integers (\type{int8}).  If a scalar variable is represented with more than $8$ bits, then the variable is subscript with the number of bits used.  For example, $a_{32}$ is a scalar represented as $32$-bit signed integer (\type{int32}).  As shown in \figref{fig:intflow}, except for quantizing floating point (\type{fp32}) data input during initialization of the training scheme, no floating point arithmetic and no further quantization is performed in NITI.

\subsection{A Block Exponentiation Scaling Scheme}

To allow the range of representable values be dynamically adjustable during training, the neural network model's activations ($\boldsymbol{a}$), gradients of activations ($\boldsymbol{e}$) and gradients of weights ($\boldsymbol{g}$) are each coupled with their own associated scaling exponent $s_a$, $s_e$ and $s_g$ respectively.  All elements of the vector share the same scaling exponent, making the scaling exponent adjustable for each layer of the model.  A scaling exponent $s$ carries a weight of $2^s$.  For example, given an activation vector $\boldsymbol{a}$ and its coupled scaling exponent $s_a$, the actual values of the model activation are $\boldsymbol{a} \cdot 2^{s_a}$.  Since the scaling exponent is shared among the entire vector and is representable as an \type{int8} value, the storage overhead is minimal.
The computation overhead of the block scaling exponentiation is also minimal. 
For instance, the scaling factor of the convolution output between $\boldsymbol{w} \cdot 2^{s_w}$ and $\boldsymbol{a} \cdot 2^{s_a}$ can be obtained by a single \type{int8} addition.

Most importantly, when compared with other works that utilize similar block scaling schemes, our training framework handles the scaling factors of $\boldsymbol{a}$, $\boldsymbol{w}$, $\boldsymbol{e}$ and $\boldsymbol{g}$ differently in order to optimize the training performance and minimize computation overhead.
In the case of activations ($\boldsymbol{a}$), $s_a$ is passed along with layer activation and is adjusted dynamically to help the results fully utilize the \type{int8} range after shift-and-round (\secref{sec:rounding}).
Furthermore, the cross entropy loss is also approximated with integer arithmetic based on the $s_a$ of the final layer.
In the case of $\boldsymbol{e}$ and $\boldsymbol{g}$, the learning rules in NITI  have been co-designed with the integer cross entropy loss computation to embed the computation of the scaling exponents $s_g$ and $s_e$ into the main algorithm.
As a result, although $s_g$ and $s_e$ are part of the underlying training logic, their computation can be optimized out in the final implementation, as indicated by the underscores in \algoref{alg:int-train}.  See \secref{sec:weight_update} and \secref{sec:int8-loss} for details.

Finally, $s_w$ remains static during the entire training.
As shown in \figref{fig:intflow}, $\boldsymbol{w}$ is not the result of a matrix multiplication.
Its bitwidth won't be expanded dramatically like $\boldsymbol{a},\boldsymbol{e}$ and $\boldsymbol{g}$. 
Hence, using shift-and-round to adjust $s_w$ dynamically is not necessary.

\subsection{Forward and Backward Pass}
At the core of NITI are the forward pass and backward pass of convolution and fully connected layers performed with integer-only arithmetic.
For convolution layers, our framework employed implicit GEneral Matrix Multiply(GEMM) to reduce convolution operations into matrix-matrix multiplications.
As a result, similar to the case of training fully connected layers, integer matrix multiplies form the dominating operation in our training framework.
With our use of \type{int8} as input, we were able to accelerate this matrix multiply in our current implementation by leveraging the newly introduced integer matrix multiply function unit in the latest GPUs that were originally designed for inference acceleration~\cite{turing}.

During the forward pass (left side of \figref{fig:intflow}),
activation of previous layer ($\boldsymbol{a}^{(l-1)}$) enters this layer as \type{int8} with a scaling factor $s_{a^{(l-1)}}$.
The results of the matrix multiply are accumulated in \type{int32} precision and must be rounded back to \type{int8} before propagating to the next layer.
This rounding operation is very important to the success of low precision training and repeatedly appears in
forward pass, backward pass and model weights update.
They are shown as shift operators in \figref{fig:intflow} as they are combined with the scaling operation in our scheme.

In theory, the output scaling factor $s_{a^{(l)}}$ is simply a sum of the $2$ input scaling factors $s_{a^{(l-1)}}$ and $s_w$.
However, to make full use of the signed $8$-bit precision to represent the integer part of the results, an additional shift operation is performed based on the \emph{effective bitwidth} of the $32$-bit matrix multiply output.
Here, effective bitwidth of an integer matrix $\boldsymbol{V}$, denoted as $B(\boldsymbol{V})$, is defined as the minimum number of bits required to fully represent the maximum value $v \in \boldsymbol{V}$.
If $b=B(\boldsymbol{V})>7$, then the $32$-bit results are shifted right by $b-7$ bit, just enough to maximize the use of the \type{int8} datatype.  The fractional part is rounded off using the pseudo stochastic rounding scheme described in \secref{sec:rounding} and the scaling factor is adjusted accordingly.

The backward pass involves propagating the error back through the integer network and computing the gradient of weights in each layer for weight update as shown in the right hand side of \figref{fig:intflow}.
The $32$-bit errors are rounded similarly to the forward pass back into $8$-bit values before being propagated to the next layer.
The $32$-bit gradients are rounded with special procedures that fuse with the weight update process as explained in \secref{sec:weight_update}.

\subsection{Weight Update}\label{sec:weight_update}
Normally with SGD, the gradient $\boldsymbol{g}^{(l)}$ of layer weights can be computed by a matrix multiplication between its activation input $\boldsymbol{a}^{(l-1)}$ and gradients of its activation output $\boldsymbol{e}^{(l)}$.
The model weights could subsequently be updated by some combinations of this gradient $\boldsymbol{g}$, the global learning rate and other heuristics to determine the amount of weight update.

However, due to the limited range of our \type{int8} weights $\boldsymbol{w}$, the task of maintaining any necessary precision is challenging, notably due to the range discrepancy between $\boldsymbol{w}$ and the desired update value.
For example, we observed empirically that even in a medium-size network, direct application of typical learning rules often resulted in
overflow or underflow in the weight update.
Consequently, most of the update values are saturated as $\pm 127$ or $0$.

Instead, we propose a learning heuristic that combines update with the rounding of the gradient for weight computation in \type{int8}.
The proposed learning heuristic draws inspiration from the Resilient backpropagation (RPROP) algorithm~\cite{Riedmiller:1993:ICNN}. 
In the original RPROP algorithm, $\boldsymbol{w}$ was updated with only the sign of $\boldsymbol{g}^{(l)}_{\texttt{32}}$.
In our case, in addition to the sign, we also utilize the effective bitwidth of $\boldsymbol{g}^{(l)}_{\texttt{32}}$ to decide the magnitude of the update.

In particular, let $b=B\left(\boldsymbol{g}^{(l)}_{\texttt{32}}\right)$ be the effective bitwidth of $\boldsymbol{g}^{(l)}_{\texttt{32}}$, then $\boldsymbol{g}^{(l)}_{\texttt{32}}$ is shifted and rounded by $b-m_u$ bits to obtain an effectively $m_u$ bits weight update value $\boldsymbol{g}^{(l)}$.
Recall that the value of $s_w$ for each layer is set during initialization and remains unchanged during training.
As a result, the proposed update heuristic essentially operates by updating $\boldsymbol{w}$ with small quantum $\boldsymbol{g}^{(l)} \cdot 2^{s_w}$.
We have evaluated different values of $m_u$ and have determined that values of $m_u$ in the range of $1$ to $5$ bits performed well in general.
We show empirically that this learning rule has comparable convergence speed
on MNIST and CIFAR10 to SGD with momentum, which is commonly used in floating point training.
See \secref{sec:mnist-cifar10-results} for experimental results.
On ImageNet, this learning rule has comparable convergence speed with SGD without momentum.

\subsection{Shifting and Rounding}\label{sec:rounding}

Mapping \type{int32} results from matrix-multiply back to \type{int8} data for downstream computation is a crucial step that has a significant influence in the final training accuracy.
To propagate activations and errors during training, a special \emph{shift and round} scheme defined in \algoref{alg:shift-round} is employed.
Using this scheme, the values in concern (i.e. $\boldsymbol{a}$ and $\boldsymbol{e}$) are first logically shifted right by an amount that is determined by their effective bitwidth $B(\boldsymbol{a})$ and $B(\boldsymbol{e})$ respectively (See lines \ref{algln:sara} and \ref{algln:sare} in Algorithm~\ref{alg:int-train}).
This shift avoids overflow and maximize the number of useful bits in the final \type{int8} representation.
In addition, the corresponding activation scaling factor $s_a$ is adjusted according to maintain correct magnitude.
\begin{algorithm}[tbp]
\caption{\textsc{ShiftAndRound}}
\label{alg:shift-round}
\SetKwInOut{Input}{Input}
\SetKwInOut{Output}{Output}
\Input{$q_{\texttt{32}}$ : $32$-bit fixed point number,\\
$s_{q_\texttt{32}}$ : the scaling factor of $q_{\texttt{32}}$,\\
$\mathit{bp}$ : Binary point}
\Output{$q$ : Rounded $8$-bit integer,\\
$s_q$ : the scaling factor of $q$}
$s_q \gets s_{q_\texttt{32}} + \mathit{bp}$\;
$q \gets $\Call{Round}{$q_{\texttt{32}}$  ,$\mathit{bp}$}\;
\end{algorithm}
Next the \type{int32} values are rounded to \type{int8}.
In the context of NITI, rounding refers to the task of mapping a $32$-bit \emph{fixed point} number $x = \langle q.f \rangle$ to a nearby $8$-bit integer $\tilde{x}$.
In this notation, $q$ and $f$ are the integer and fraction parts of $x$ respectively, and the binary point is located at position $bp$, which is equal to the bitwidth of $f$.

As discussed in \cite{stochastic-rounding}, the use of stochastic rounding with zero rounding bias is crucial to the success of training neural networks with low precision.
Stochastic rounding $\Call{Rs}{\cdot}$ can be defined as:
\[
\Call{Rs}{\langle q.f \rangle} = \begin{cases}
    q & \text{with probability } 1-f\cdot2^{-bp} \\
    q+1 & \text{with probability }  f\cdot2^{-bp}\\
   \end{cases}
\]

\begin{figure}[htb]
    \centering
    \includegraphics[width=\linewidth]{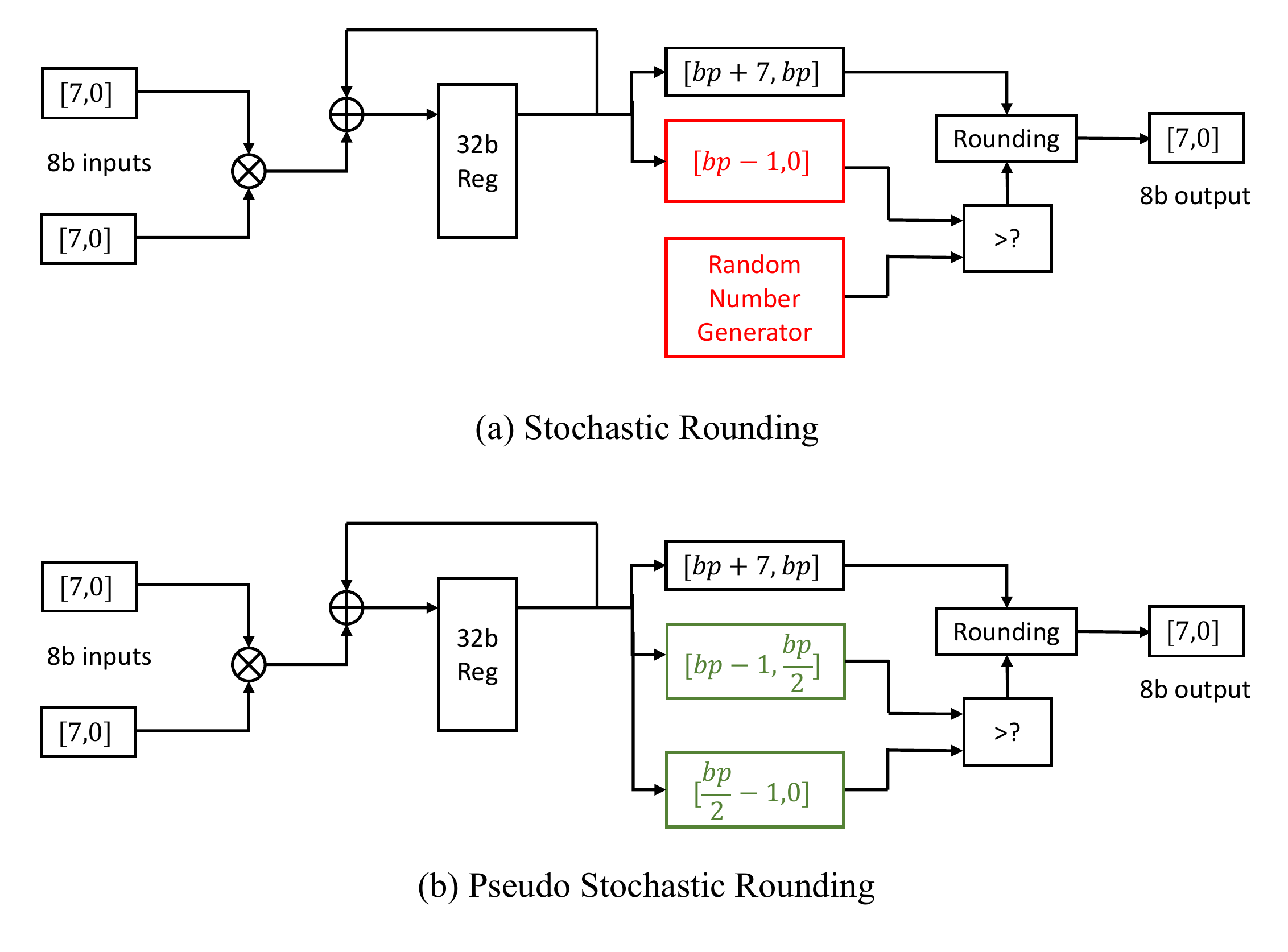}
    \caption{Rounding Schemes Hardware Implementation}
    \label{fig:round_hw}
\end{figure}

A typical implementation of the stochastic rounding function would therefore compute the rounding probability by comparing $f$ with a random number $r$, as shown in \figref{fig:round_hw}.

Instead of relying on an external random number source to produce $r$, we propose a hardware-efficient \emph{pseudo stochastic rounding} algorithm that generates in-situ pseudo random numbers using the additional bits from the $32$-bit input.
As shown in \algoref{alg:psto}, the proposed scheme operates in ways similar to the original stochastic rounding function, except the stochastic rounding decision is now based on comparing values between different portions of $f$, the fractional parts of the input.
Our current implementation divides the fractional bits into $2$ halves for comparison.
The more significant (top) half of $f$ is essentially a truncated version of $f$, and the less significant (bottom) half of $f$ now serves as a pseudo random number.
As shown in \secref{sec:rounding_compare}, the proposed pseudo stochastic rounding scheme is able to support training with comparable accuracy as the original stochastic rounding scheme.

\begin{algorithm}[tbp]
\caption{Pseudo Stochastic Rounding}
\label{alg:psto}
 \SetKwInOut{Input}{Input}
 \SetKwInOut{Output}{Output}
\Input{$q_{\texttt{32}}$: $32$-bit fixed point number,  
$\mathit{bp}$: Binary point}
\Output{Rounded $8$-bit integer $q$}
\tcc{Extract bits for integer ($q$) and fractional ($f$) parts from $|q_{\texttt{32}}|$}
$q,\;f \, \gets \, |q_{\texttt{32}}|[bp+7, bp],\; |q_{\texttt{32}}|[bp-1, 0]$\;
\If{$bp$ \textup{is odd}}
{
$f \gets f \gg 1$\tcc{right shift by 1 bit}
$bp \gets bp-1$\;
}
\If{$f[bp-1,\frac{bp}{2}] > f[\frac{bp}{2}-1,0]$}
{
$q \gets q+1$
}
$q \gets q \cdot \sgn(q_{\texttt{32}})$

\end{algorithm}

\begin{figure}
\centering
\subfigure[Activation $\boldsymbol{a}$]{\includegraphics[width=\linewidth]{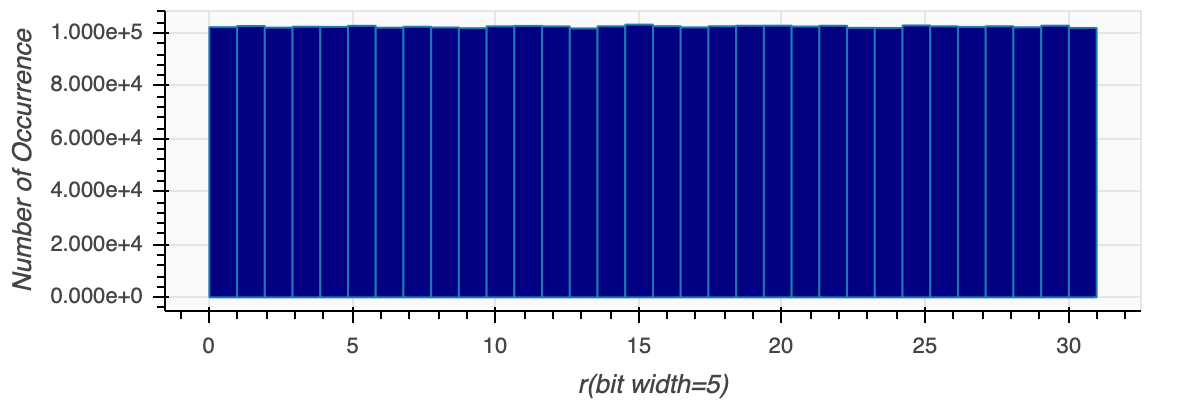}\label{fig:act_prng}}
\subfigure[Error $\boldsymbol{e}$]{\includegraphics[width=\linewidth]{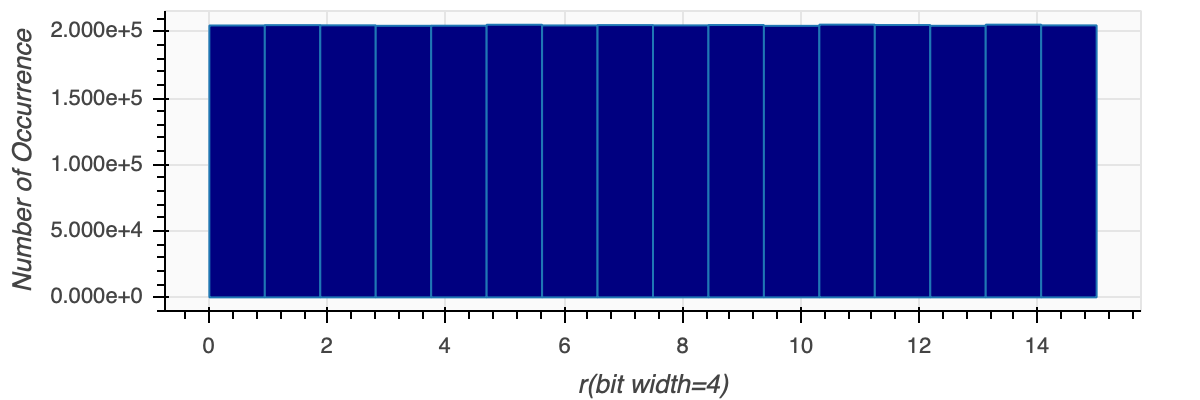}\label{fig:err_prng}}
\subfigure[Gradient $\boldsymbol{g}$]{\includegraphics[width=\linewidth]{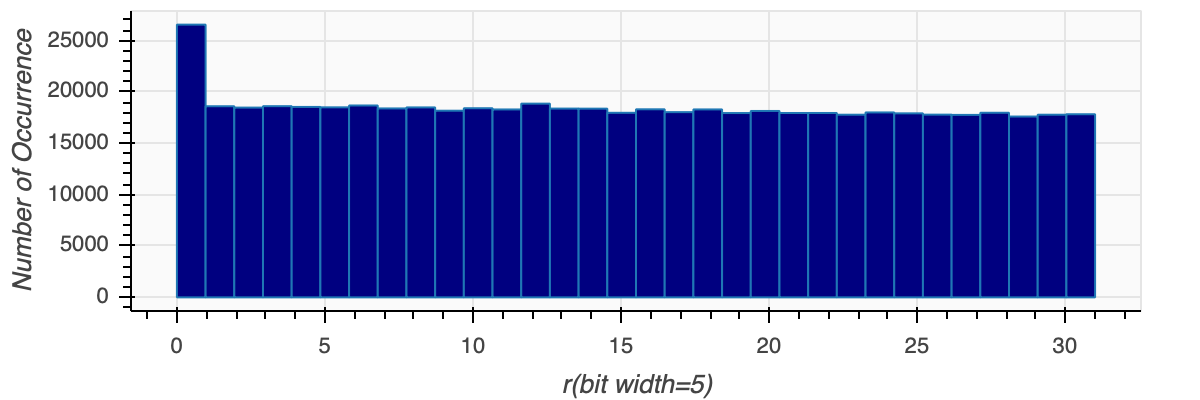}\label{fig:grad_prng}}
\caption{Histogram of pseudo random number ($r$) used for rounding different variables during training of one convolution layer.}
\label{fig:hist_prng}
\end{figure}

\figref{fig:hist_prng} shows histograms of the pseudo random number used for rounding $\boldsymbol{a}$, $\boldsymbol{e}$, $\boldsymbol{g}$ in one convolution layer during the course of training.
The figures show that 
our scheme is able to produce overall random numbers with near uniform distribution.
Similar distribution in other convolution and fully connected layers can also be made.
\subsection{Integer Cross Entropy Loss}\label{sec:int8-loss}
After computing the final layer activations, $\boldsymbol{a} \cdot 2^{s_a}$,
NITI computes the predicted probability of each class for the input image using a softmax layer.
Assume we have $N$ classes with $i \in \{1 \ldots N\}$, then the predicted probability of class $i$ is
\[
\hat{y_i} = \frac{e^{a_i\cdot 2^{s_a}}} {C}
\]
where $C=\sum_{i=1}^{N} {e^{a_i\cdot 2^{s_a}}}$.

With the predicted probability $\boldsymbol{\hat y}$ and the target one hot probability $\boldsymbol{y}$ obtained from the labels, the backpropagation process can be initiated by computing the gradient $\boldsymbol{e}$ of cross entropy loss, $E$.  Cross-entropy loss is defined as \[E = -\sum_{i=1}^{N} {y_i\ln(\hat{y_i})}\] and its partial derivative can be computed as:

\begin{align}\label{eq:cross-entropy}
e_i &= \frac{\partial E}{\partial (a_i\cdot 2^{s_a})} \nonumber\\
&= \hat{y_i}-y_i \nonumber\\
&= \frac{1}{C}\left(e^{a_i\cdot 2^{s_a}} - y_iC\right)\quad\quad i\in \{1 \ldots N\}
\end{align}

Consider \eqref{eq:cross-entropy}, although the factor $\frac{1}{C}$ is difficult to compute using integer arithmetic, note that mathematically it is a constant (independent of $i$) that affects only the scaling of $e_i$.  In other words, it only affects the scaling factor ($s_e$) of $\boldsymbol{e}$ but not the relative values among the elements of $\boldsymbol{e}$.
Now, as shown in \secref{sec:weight_update}, we have designed our learning rule in such a way that it doesn't depend on $s_g$. 
Consequently, the actual value of $s_e$ also has no effect on the final results and thus the computation of $\frac{1}{C}$ can be avoided entirely.
The main challenge of implementing \eqref{eq:cross-entropy} with integer arithmetic is therefore to approximate the term $t_i = e^{a_i\cdot 2^{s_a}}$ accurately with limited precision.
We address this challenge by considering $2$ cases.

When $s_a\le-7$, we know $|a_i\cdot 2^{s_a}| < 1$ because, as an $8$-bit integer, $a_i \le 127$.  We can therefore approximate $t_i$ via its Taylor expansion:
\begin{align}
    t_i &= e^{a_i \cdot 2^{s_a}} \nonumber\\
        &\approx 1 + a_i \cdot 2^{s_a}+ \frac{1}{2} a_i^2  \cdot 2^{2s^a}\label{eq:sa_small}
\end{align}

When $s_a>-7$, we rearrange the term with a base-$2$ approximation as follows:
\begin{align}
t_i &= e^{a_i \cdot 2^{s_a}}\nonumber\\
&=2^{(\log_2 e) \cdot a_i \cdot 2^{s_a}}\label{eq:sa_large}
\end{align}

Note that in both cases, $2^x$ corresponds to a shift operation in integer fixed point representations if the exponent $x$ is an integer.
In the case of \eqref{eq:sa_small}, since $s_a$ and $2s_a$ are both integers, $t_i$ can be computed by using simple integer add and shift.
In the case of \eqref{eq:sa_large}, the value of $t_i$ is a power-of-$2$, but the exponent value is not necessarily an integer.

To efficiently approximate the exponent in \eqref{eq:sa_large} as an integer, denote the exponent in \eqref{eq:sa_large} as $x_i = (\log_2 e) \cdot a_i \cdot 2^{s_a}$.  Now,
\[
(\log_2 e) = 1.44269 \approx 47274\cdot 2^{-15}
\]
is a constant, which allows us to approximate $x_i$ as
\begin{align}
    x_i &\approx (47274 \cdot 2^{-15}) \cdot a_i \cdot 2^{s_a}\nonumber\\
    &= (47274 a_i) \cdot 2^{s_a - 15} \label{eq:xi_approx1}
\end{align}
In \eqref{eq:xi_approx1}, the constant \num{47274} $=$ \texttt{0xB8A8} in hex requires $16$ bits to represent, while $a_i$ is an $8$-bit integer, making their product a $24$ bit integer.
We observe empirically that $s_a<0$ in most cases.
Therefore, $x_i$ can be approximated by shifting the value $47274 a_i$ \emph{right} by $|s_a|+15$ followed by truncation of any resulting fractional bits.  Denote this integer approximation of $x_i$ as $\hat{x}_i$.

Although $t_i$ is now ready to be approximated by $2^{\hat{x}_i}$ with a shift operation, $2^{\hat{x}_i}$ can still be a very long integer if $\hat{x}_i$ is large.
We apply an offset $p$ to $\hat{x}_i$, so $t_i=2^p\cdot2^{\hat{x}_i-p}$.
The term $2^p$ can be absorbed into $\frac{1}{C}$ and its computation can also be avoided.
Note that for any value $\hat{x}_i<p$, after the shift operation, it's at least $2^{max(\boldsymbol{\hat{x}})-p}$ times smaller than the maximum value in $t_i$ which is $2^{max(\boldsymbol{\hat{x}})}$.
We choose $p$ be the smallest value of $\hat{x}_i$ larger than $\max(\boldsymbol{\hat{x}}) - 10$, 
so for any value $\hat{x}_i<p$, after the shift operation, it is at least $2^{10}$ smaller than $2^{max(\boldsymbol{\hat{x}})}$.
These small values don't influence $e_i$ after rounding back to \type{int8}, so they can be safely clipped away.
Denote $\tilde{x}_i = \max(0, \hat{x}_i - p)$ to be the values after applying the offset and clipping to $\hat{x}_i$.
Finally, $t_i = 2^{x_i}$ is approximated as $2^p \cdot 2^{\tilde{x}_i}$.
The computation of $2^p$ can be avoided as mentioned before.
$2^{\tilde{x}_i}$ can be computed by $1 \ll \tilde{x}_i$ and its maximum value is $2^{10}$.
After substituting $t_i$ back to Equation~\eqref{eq:cross-entropy},
$e_i$ can be represented by a $10$-bit integer times an scaling factor for the case $y_i=0$.
For the case $y_i=1$, we need to add all $2^{\tilde{x}_i}$. 
In this addition step, we use the same bitwidth as the intermediate accumulation of \type{int8} matrix multiplication.

The error tensor $\boldsymbol{e}$ in (\ref{eq:cross-entropy}) eventually rounded stochastically back to $8$ bits before being used in back propagation. 
For a dataset with \num{1000} classes like ImageNet, stochastic rounding is necessary to avoid round bias in $e_i$.  Despite the crude approximation in some cases, we find the accuracy of the resulting network competitive to related works that depend on floating-point implementations.
\section{Experimental Results}\label{sec:experiment}
In this section, we first evaluate the key factors which influence the performance of trained networks by NITI framework including weights initialization schemes and different rounding schemes for $\boldsymbol{g}$ .
To do so, we used CIFAR10~\cite{cifar10} datasets with a network with 7 weight layers similar to VGG~\cite{vgg}(referred as VGG-small-7 in this section). 
Detailed network configuration is shown in the first column of table \ref{table:cifar10-results}.
Learning from the above evaluation, we optimized our NITI framework using the proposed rounding schemes and weight initialization scheme.
Then, we compare the performance of NITI framework with its floating point counterpart trained by SGD optimizer over MNIST~\cite{mnist}, CIFAR10, and ImageNet~\cite{deng2009imagenet} datasets.

\subsection{Weight Initialization}
As mentioned in \secref{sec:weight_update}, the training updates are performed directly on \type{int8} weights, which have significantly less dynamic range than \type{fp32} weights.
Related mixed precision training works\cite{wageubn,8bit-integer-training} usually initialize the weights with a zero-mean normal distribution, just like the floating-point training. 
Such initialization is rather sparse in the sense that most of the values are small around zero.
We try to compensate the limited weight update precision by a denser weight initialization using uniform distribution.
As for the scaling factor $s_w^{(l)}$ associated with the integer weights $\boldsymbol{w}^{(l)}$, it is determined by similar rules used in ~\cite{He:2015, Glorot:2010}.
The idea is that the scaling factor should be initialized in such a way to avoid inputs being amplified.

The task of training VGG-small-7 on CIFAR10 was chosen to demonstrate the efficiency of two weight initialization schemes: weights with uniform(dense) distribution and normal(sparse) distribution.
The average top-$1$ validation accuracy of 5 training experiments with each initialization scheme is shown in the \tabref{table:compare-init-scheme}.
As shown in the table, we observed that using uniformly distributed(dense) network weight initialization generally results in better validation accuracy.

\begin{table}[tb]
    \begin{center} 
    \caption{Average top-$1$ validation accuracy(5 runs) of training VGG-small-7 on CIFAR10 with different weight initialization schemes. 
            The rounding scheme for $\boldsymbol{g}$ was stochastic rounding.
            The rounding scheme for $\boldsymbol{a,e}$ was round-to-nearest.
            $m_u$ was initiated by $5$ bits.
            It was dropped to $4$ bits at the $100^{th}$ and $3$ bits at the $150^{th}$ epoch.
            The training was stopped at the end of the $200^{th}$ epoch.} 
    \label{table:compare-init-scheme}
        \begin{tabular}{cc} 
            \toprule
            Distribution& Validation Accuracy(\%)\\
            \midrule
            Normal & 87.3 $\pm$ 0.3\\
            Uniform & 90.8 $\pm$ 0.2\\
            \bottomrule
        \end{tabular} 
    \end{center}
\end{table}

\begin{figure}[htb]
\caption{Weight distribution before and after training using normal(up) and uniform initialization(down).}
    \label{fig:hist-compare}
    \begin{center}
       \includegraphics[width=1.0\linewidth]{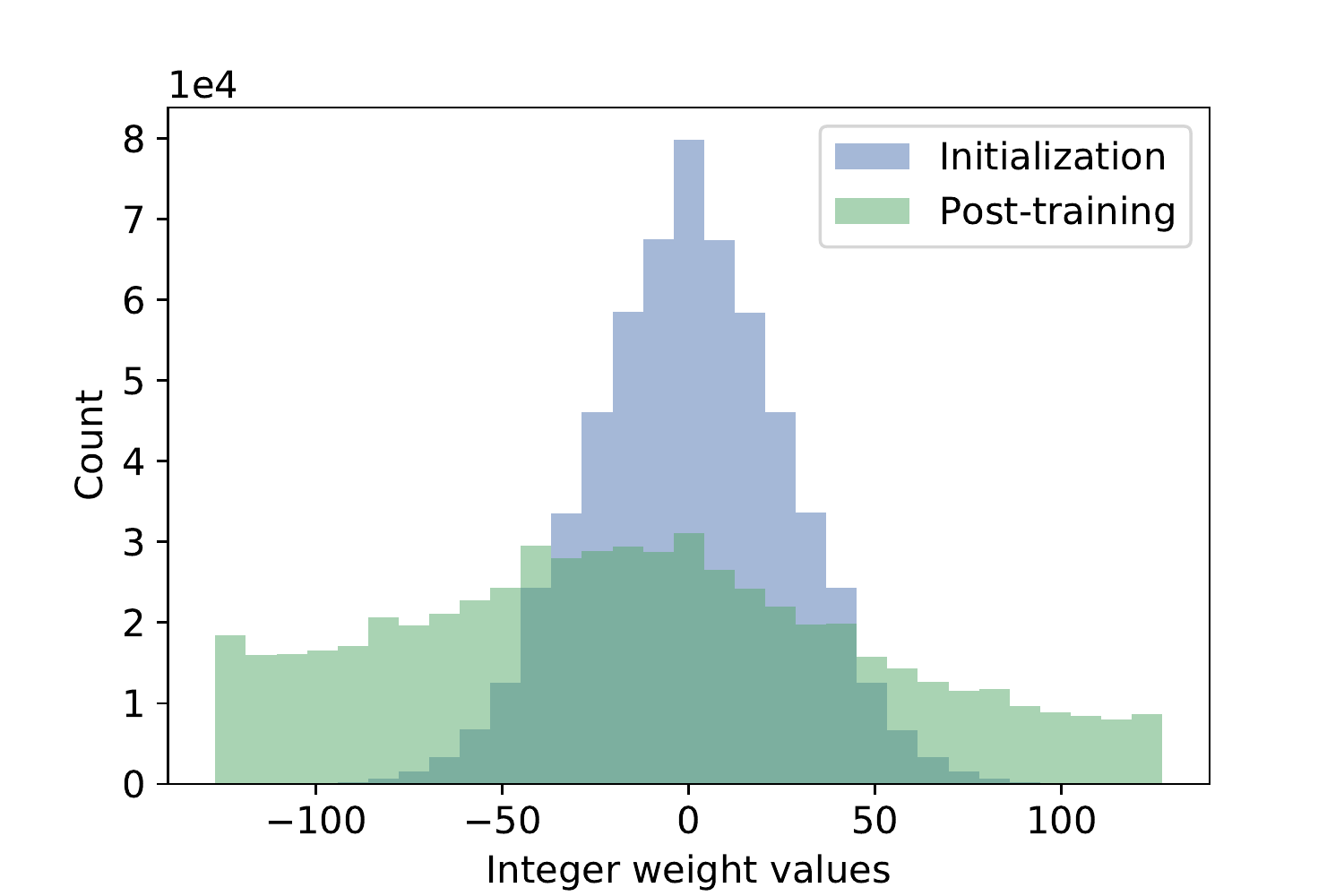}
       \includegraphics[width=1.0\linewidth]{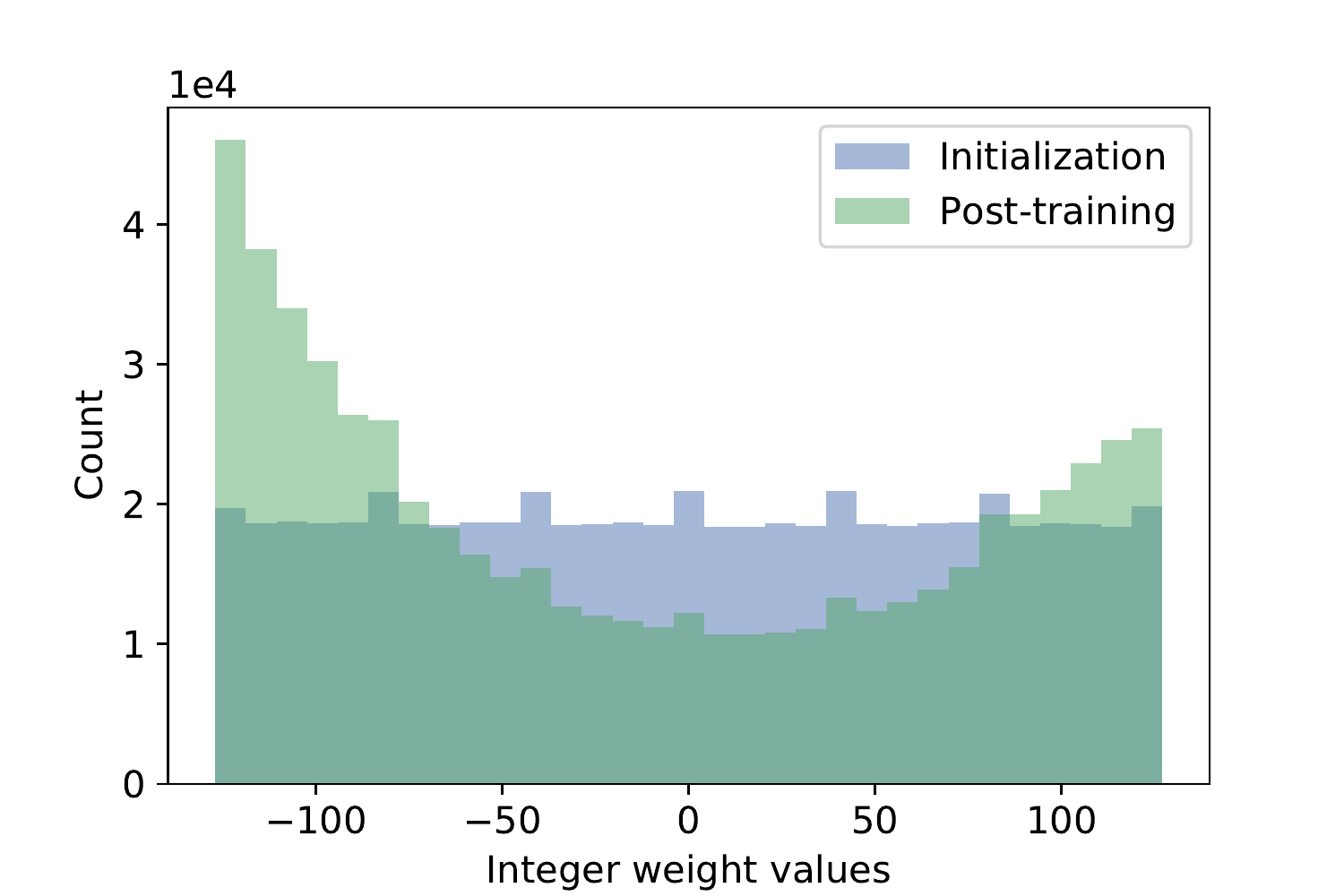}
    \end{center}
\end{figure}

Figure \ref{fig:hist-compare} compared the weight distribution of a typical convolution layer inside VGG-small-7 before and after the training under the two initialization schemes.
The learning heuristic introduced in Section 3.2 adjusts weights to have fewer zeros with proper initialization.
We also observed similar weight distribution while using NITI to train networks shown in \tabref{table:cifar10-results} on CIFAR10 and AlexNet on the ImageNet dataset.

We hypothesize that floating-point training has more redundancies in the weight precision, so they tend to have more zero weights.
However, if we accumulate the weights in int8 precision directly, 
the final trained model compensates for the limited precision with more non-zeros weights.
Some works\cite{Yang_2020_CVPR,srivastava2019joint} demonstrate the trade-off between sparsity and weights bit width in the sense of inference acceleration.
However, such a compensation mechanism has never been demonstrated in previous low precision training works.
We also observe that weights in the first layer are almost zero-free, both in CIFAR10 and ImageNet experiments.
We think that is why related low precisoin training works\cite{wage,wageubn} needs to keep the first layer as floating-point precision to prevent accuracy loss.

\subsection{Rounding Scheme}\label{sec:rounding_compare}
As stochastic rounding is adopted by most recent low precision training implementations~\cite{8bit-integer-training,wage,fxpnet-2017}, 
we use it as the baseline for evaluating our pseudo stochastic rounding scheme.
We also include results from straight-forward round-to-nearest for comparison.
\tabref{table:rounding-scheme} shows the averge top-$1$ validation accuracy of 5 experiments for training \type{int8} VGG-small-7 on CIFAR10 dataset with different rounding schemes. 
Our pseudo stochastic rounding technique works as well as baseline stochastic rounding while both significantly outperform round-to-nearest method in terms of final best validation accuracy.

\begin{table}[tb]
    \begin{center} 
    \caption{Average top-$1$ validation accuracy(5 runs) of training VGG-small-7 on CIFAR10 with different rounding schemes for $\boldsymbol{g}$. 
            The rounding scheme for $\boldsymbol{a,e}$ was round-to-nearest.
            Weights were initiated using uniform distribution.
            $m_u$ schedule was the same as \tabref{table:compare-init-scheme}.}
    \label{table:rounding-scheme}
        \begin{tabular}{cc} 
            \toprule
            Rounding schemes & Validation Accuracy(\%)\\
            \midrule
            Round-to-nearest & 89.5 $\pm$ 0.1\\
            Stochastic & 90.8 $\pm$ 0.2\\
            Pseudo Stochastic & 90.7 $\pm$ 0.1\\
            \bottomrule
        \end{tabular} 
    \end{center}
\end{table}

\subsection{Accuracy Results on MNIST and CIFAR10 datasets}\label{sec:mnist-cifar10-results}
We trained LeNet~\cite{lenet} and VGG-small networks respectively over MNIST and CIFAR10 datasets, using both optimized NITI framework and the baseline \type{fp32} method. 
The \type{fp32} baseline technique uses SGD with momentum $0.9$. 
For the case of LeNet model, we fixed the learning rate in \type{fp32} training to $0.01$ and $m_u$ to $3$ bits in \type{int8} training.
Both \type{fp32} and \type{int8} training processes were stopped at the end of $20^{th}$ epoch. 
In the case of VGG-small, we initiated the learning rate by $0.01$ in \type{fp32} training. It was dropped to $0.001$ at $100^{th}$ epoch and $0.0001$ at $150^{th}$ epoch.
In \type{int8} training, we initiated $m_u$ by $5$ bits.
It was dropped to $4$ bits at $100^{th}$ epoch and $3$ bits at $150^{th}$ epoch.
The rounding scheme for $\boldsymbol{a,e}$ was round-to-nearest.
The rounding scheme for $\boldsymbol{g}$ was pseudo-stochastic rounding.
Weights were initiated using uniform distribution.
Both \type{fp32} and \type{int8} training processes were stopped at the end of $200^{th}$ epoch when the training accuracy stopped improving. 

\begin{figure}[tb]
    \centering
    \includegraphics[width=\linewidth]{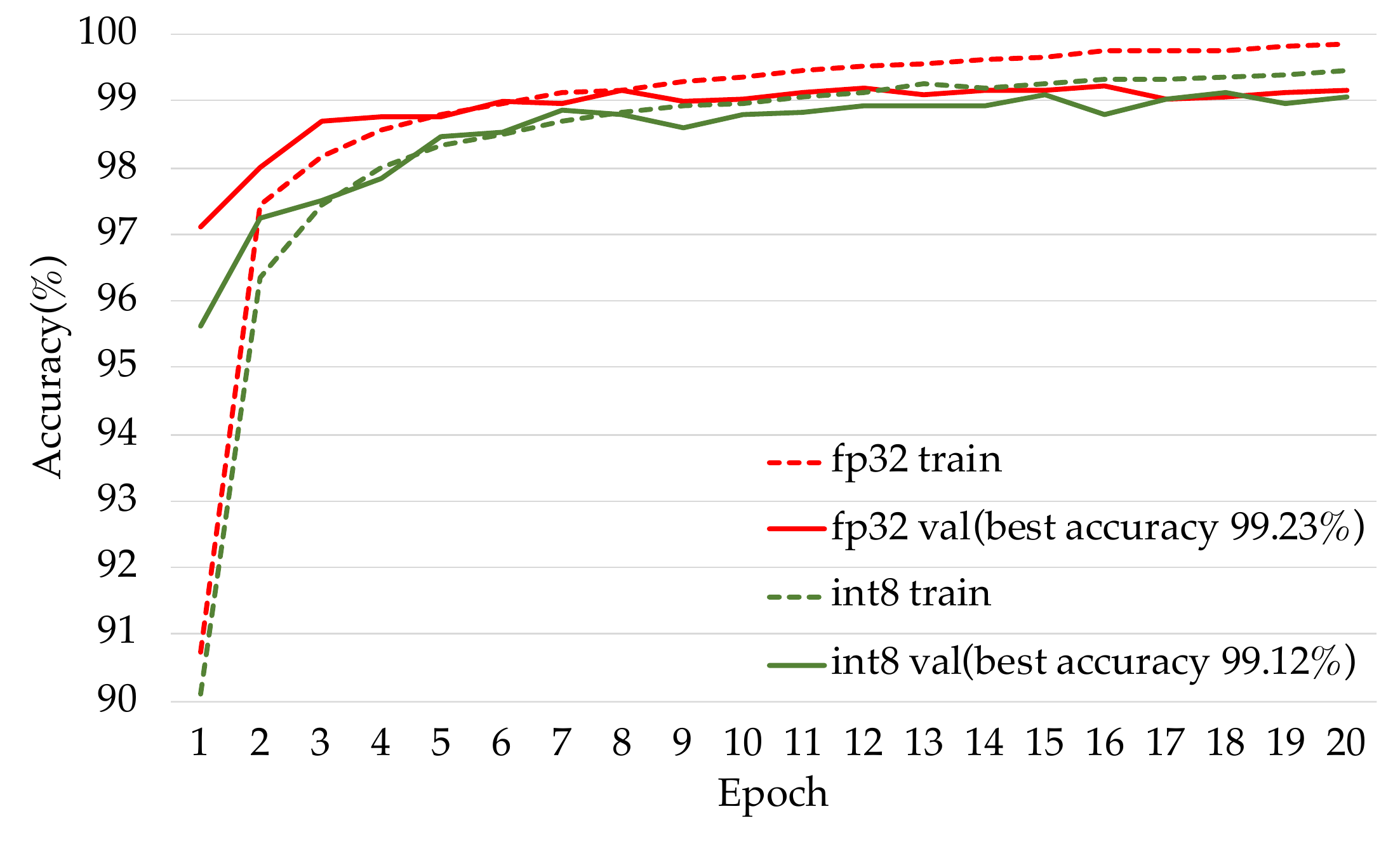}
    \caption{Training performance comparing NITI (\type{int8}) and baseline \type{fp32} implementation on MNIST}
    \label{fig:mnist_int8_vs_fp32}
\end{figure}

\begin{figure}[tb]
    \centering
    \includegraphics[width=\linewidth]{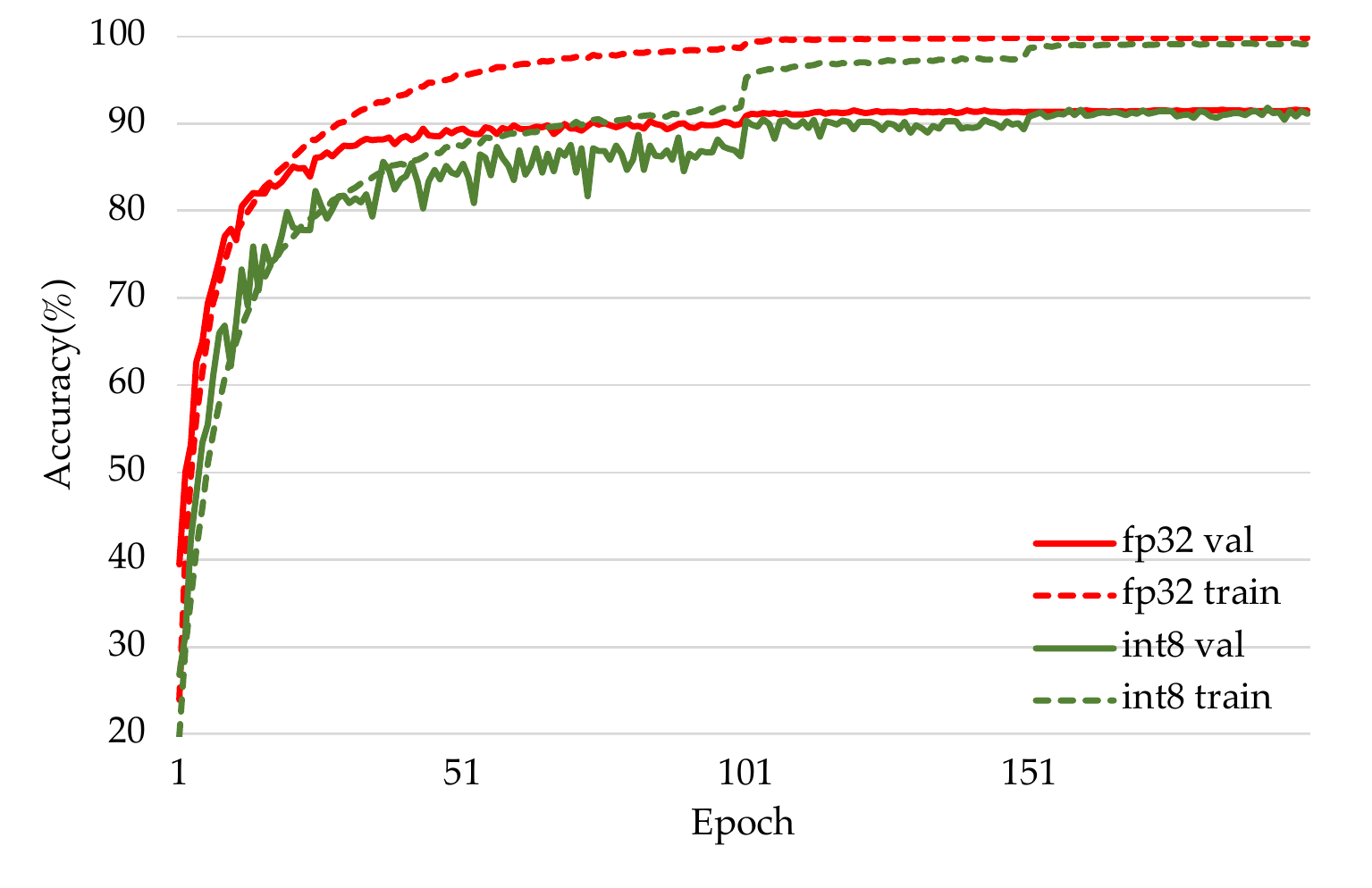}
    \caption{Training VGG-small (9 weight layers) comparing NITI (\type{int8}) and baseline \type{fp32} implementation on CIFAR10}
    \label{fig:vgg_small_int8_vs_fp32}
\end{figure}

As demonstrated in \figref{fig:mnist_int8_vs_fp32}, NITI trains LeNet model, which is a relatively small network, with almost identical performance comparing to the \type{fp32} baseline. 
In the case of training VGG-small-7 over CIFAR10, which is a tougher task, NITI was able to successfully train the network with negligible descend, achieving a top-$1$ validation accuracy of \SI{90.7}{\percent}. 
\tabref{table:accuracy-results}
summarizes the validation accuracy of our framework on CIFAR10 while increasing network depth. 
As observed, deeper network produced the better accuracy by NITI.
NITI even produced slightly better generalization performance than the \type{fp32} baseline for VGG-small-8 and VGG-small-9(\figref{fig:vgg_small_int8_vs_fp32}).

\begin{table} 
    \begin{center} 
    \caption{VGG-small network configurations}
    \label{table:cifar10-results}
        \begin{tabular}{|c|c|c|} 
            \hline
            7 weight layers & 8 weight layers & 9 weight layers\\
            \hline
            conv3-128 & conv3-128 & conv3-128\\
            conv3-128 & conv3-128 & conv3-128\\
                      & conv3-128 & conv3-128\\
            \hline
            \multicolumn{3}{|c|}{maxpool}\\
            \hline
            conv3-256 & conv3-256 & conv3-256\\
            conv3-256 & conv3-256 & conv3-256\\
                      &           & conv3-256\\
            \hline
            \multicolumn{3}{|c|}{maxpool}\\
            \hline
            conv3-512& conv3-512& conv3-512\\
            conv3-512& conv3-512& conv3-512\\
            \hline
            \multicolumn{3}{|c|}{maxpool + dropout + FC}\\
            \hline
        \end{tabular} 
    \end{center}
    
\end{table}

\subsection{Accuracy Results on ImageNet dataset}

\begin{table*}[htbp]
  \centering
  \caption{Comparison to related works}
  \label{table:accuracy-results}
    \begin{tabular}{cccccccccccc}
    \toprule
    Dataset & Model & Size & Related works & {$\boldsymbol{w}(infer)$} & {$\boldsymbol{w}(acc)$} & {$\boldsymbol{a}$}     & {$\boldsymbol{g}$}    & {$\boldsymbol{e}$}  & Loss  & Accuracy &  \XX{Memory}\\
    \midrule
    MNIST & LeNet 5 & 6.87K & fp32 baseline & fp32  & fp32  & fp32  & fp32  & fp32  & fp32  & $99.2\%$ &2019 MiB \\
    MNIST & LeNet 5 & 6.87K & NITI (\textit{this work}) & 8     & 8     & 8     & 5     & 8     & integer & $99.1\%$ &1538 MiB\\
    CIFAR10 & VGG-small 7 & 4.66M & fp32 baseline & fp32  & fp32  & fp32  & fp32  & fp32  & fp32  & $91.0\pm0.1\%$ &5825 MiB\\
    CIFAR10 & VGG-small 7 & 4.66M & NITI (\textit{this work}) &  8    & 8     & 8     & 5     & 8 & integer & $90.7\pm0.1\%$ & 3169 MiB\\
    CIFAR10 & VGG-small 8 & 4.72M & fp32 baseline & fp32  & fp32  & fp32  & fp32  & fp32  & fp32  & $91.3\pm0.2\%$ &5957 MiB\\
    CIFAR10 & VGG-small 8 & 4.72M & NITI (\textit{this work}) &  8    & 8     & 8     & 5     & 8 & integer  & $91.5\pm0.2\%$ &3301 MiB \\
    CIFAR10 & VGG-small 9 & 5.39M & fp32 baseline & fp32  & fp32  & fp32  & fp32  & fp32  & fp32  & $91.5\pm0.1\%$ &6033 MiB\\
    CIFAR10 & VGG-small 9 & 5.39M & NITI (\textit{this work}) & 8     & 8     & 8     & 5     & 8     & integer & $91.8\pm0.2\%$ &3303 MiB\\
    CIFAR10 & VGG-like & 9.29M & FxpNet\cite{fxpnet-2017} & 1    & 12     & 1     & 12    & 12    & fp32  & $88.5\%$ & - \\
    ImageNet & AlexNet & 62.38M & fp32 baseline & fp32  & fp32  & fp32  & fp32  & fp32  & fp32  & $49.0\%$ &7389 MiB\\
    ImageNet & AlexNet & 62.38M & Banner~\etal~\cite{8bit-integer-training} & 8     & 32    & 8     & 8     & 8     & fp32  & $48.9\%$ &9317 MiB\\
    ImageNet & AlexNet & 62.38M & WAGE\cite{wage}  & 2     & 8(32) & 8(32) & 8(32) & 8(32) & fp32  & $48.4\%$ & - \\
    ImageNet & AlexNet & 62.38M & NITI (\textit{this work}) & 8     & 8     & 8     & 5     & 8     & integer & $48.3\%$ &6369 MiB \\
    ImageNet & AlexNet & 62.38M & DFP-16\cite{mix-int16}  & 16    & fp32  & 16    & fp32  & 16    & fp32  & $56.9\%$ & - \\
    \bottomrule
    \multicolumn{12}{l}{(1)The VGG-small series results on CIFAR10 are the average top-$1$ validation accuracy of 5 runs.}\\
    \multicolumn{12}{l}{(2)The "Size" column lists the total number of parameters in the model.}\\
    \multicolumn{12}{l}{(3)The "Memory" column lists memory footprints of training on RTX 3090 GPU. Some data are missing("-") because codes are not available.}\\  
    \end{tabular}%
  \label{tab:addlabel}%
\end{table*}%

\tabref{table:accuracy-results} tabulates the results of several related integer-training works on different datasets.
The \type{fp32} baseline AlexNet were trained using vanilla SGD without momentum, weight decay, dropout, and batch normalization to match with the training setup of WAGE \cite{wage}.
Initial learning rate was $0.01$ and dropped to $10^{-3}$ and ${10^{-4}}$ at epoch $60$ and $65$ respectively. 
Total training epochs were $70$.
Since the original publication of \cite{8bit-integer-training} did not include AlexNet results, we obtained the comparison results by using their released code.

As mentioned in \secref{sec:related}, both \cite{8bit-integer-training} and \cite{wage} have maintained some degrees of computation using floating point arithmetic during their training process.
The result of \cite{wage} further excluded cross entropy loss and the last layer from quantization to avoid performance degradation.
\cite{mix-int16} preserved even more floating point computations in the training. Their accuracy on ImageNet is significantly higher than the comparisons because momentum, weight decay and batch normalization are all computed with floating point precision.
Being the only work that performed network training exclusively with integer arithmetic, 
NITI achieves similar accuracy on the task of training integer AlexNet compared with state-of-the-art integer training frameworks.
We plan to test the NITI algorithm on more network architectures in the future.
\section{Acceleration Results}\label{sec:acceleration}
To study the benefits of using \type{int8} operations in accelerating neural network training with NITI, two acceleration strategies have been explored.  First, we explored the use of native \type{int8} matrix multiplication in modern GPUs to accelerate the computational bottleneck of NITI.  Then, an FPGA accelerator that performed the SGD backpropagation algorithm for a fully connected layer was implemented  to study the resource and speed advantage of \type{int8} computations over their floating-point counterparts.

\subsection{Tensor Cores Acceleration in GPU}\label{sec:int8-impl}
Recognizing the importance of \type{int8} operations in accelerating deep quantized neural network inference, many modern GPUs and AI accelerators have been introducing compute engines that are dedicated to these low precision integer operations.  Since NITI relies on \type{int8} computation natively, here we evaluate how these accelerators originally designed for inference acceleration may be used to accelerate neural network training as well.

In particular, we utilized the tensor cores introduced in modern nvidia GPUs to accelerate the most time-consuming part in the training process, namely, the computation of $\boldsymbol{a}$, $\boldsymbol{e}$, and $\boldsymbol{g}$ of convolution layers as shown in \figref{fig:intflow}. 
Tensor cores in nvidia GPUs are designed to perform \type{int8} matrix multiplications with high efficiency.

The use of \type{int8} can lead to $4$ times improvement in the peak computation throughput of tensor cores 
in NVIDIA's Ampere architecture\cite{a100} when compared to \type{fp32} operations.

Accelerating the computation of $\boldsymbol{a}$ with \type{int8} precision on tensor cores was well supported on commonly used DNN training frameworks like PyTorch\cite{NEURIPS2019_9015} and TensorFlow\cite{tensorflow2015-whitepaper}. 
However, the current API (cuDNN v8.2\cite{chetlur2014cudnn}) for tensor core operations does not provide native support for computing $\boldsymbol{e}$ and $\boldsymbol{g}$ with \type{int8} precision directly.
Although the underlying computations of $\boldsymbol{e}$ and $\boldsymbol{g}$ are matrix multiplications similar to the computation of $\boldsymbol{a}$ as shown in \figref{fig:intflow},
the reduced precision imposes great challenges for maintaining the training accuracy.
With our proposed training framework, taking the speed advantage of \type{int8} precision became feasible.

\tabref{tab:aeg-speedup}
shows the speedup of  NITI's $\boldsymbol{a, e, g}$ computation under different convolution input size conditions on two different NVIDIA GPUs while compared with their equivalent floating-point implementations. 
For $\boldsymbol{a}$ and $\boldsymbol{e}$ computations, the use of \type{int8} matrix multiply provided up to $6.76\times$ and $8.57\times$ speedup on the 2080ti GPU respectively.  The performance of both \type{int8} and \type{fp32} operations improved with the newer 3090 GPU, with the gap between \type{fp32} and \type{int8} reduced, likely due to the highly optimized routine employed in the \type{fp32} implementations.

As for $\boldsymbol{g}$ computation, the speedup of \type{int8} over \type{fp32} is less significant.
$\boldsymbol{g}$ was computed by convolving the rearranged $\boldsymbol{a}^{(l-1)}$ with the rearranged $\boldsymbol{e}^{(l)}$.
The kernel (rearranged $\boldsymbol{e}^{(l)}$) used in this convolution was very different from normal inference convolutions.
In the cases shown in 
\tabref{tab:aeg-speedup},
its size varied from $224 \times 224$ to $14 \times 14$. 
Currently, we only optimized our implementation to handle small convolution kernels used in mainstream CNNs, like $3 \times 3$, $5 \times 5$ and $7 \times 7$.
Different kernel sizes require different strategies of mapping to matrix multiplication tiles of tensor cores.
We plan to optimize this part in the future. 

The last column of \tabref{table:accuracy-results} summarizes the GPU memory footprint of different training experiments. 
The evaluations were conducted using PyTorch 1.8.0 and CUDA 11.1.
The training batch sizes used on MNIST, CIFAR10, and ImageNet were 256, 256, and 512, respectively. 
Comparing with highly optimized \type{fp32} baseline, our preliminary GPU implementation of NITI reduced the end-to-end training memory footprint by up to $1.31\times, 1.84\times, 1.16\times$ on MNIST, CIFAR10 and ImageNet, respectively. \cite{8bit-integer-training} consumed more GPU memory during training, as the low precision arithmetic were emulated through \type{fp32} operations. 
The memory footprint of \cite{wage, fxpnet-2017} were not reported in \tabref{table:accuracy-results} because related training codes are not available.
We expect \cite{wage, fxpnet-2017} will consumed more GPU memory than \type{fp32} baseline due to their emulation-based implementation like\cite{8bit-integer-training}.  

An open-source implementation of NITI can be found in \cite{nitisource}, 
which includes an implementation of native \type{int8} backward pass using tensor cores that can be integrated to the PyTorch training framework as a CUDA extension.

\begin{table}[htbp]
\setlength{\tabcolsep}{5pt}
  \centering
  \caption{Speedup of convolution layers on GPUs using \type{int8} compared with using \type{fp32}.All the convolution layers had the same configuration(batch size 64, 64 input channels, 128 output channels, $3 \times 3$ kernel, stride 1, and padding 1) except the input size.}
    \begin{tabular}{ccccccc}
    \toprule
    \multicolumn{7}{c}{Computing {$\boldsymbol{a}$}} \\
    \midrule
    GPU   & \multicolumn{3}{c}{RTX 2080 Ti} & \multicolumn{3}{c}{RTX 3090} \\
    input  & int8(ms) & fp32(ms) & speedup & int8(ms) & fp32(ms) & speedup \\
    224   & 11.624 & 51.092 & 4.40  & 3.928 & 13.033 & 3.32 \\
    112   & 3.069 & 13.301 & 4.33  & 1.006 & 3.214 & 3.19 \\
    56    & 1.166 & 3.821 & 3.28  & 0.289 & 1.125 & 3.90 \\
    28    & 0.367 & 1.425 & 3.88  & 0.089 & 0.311 & 3.51 \\
    14    & 0.101 & 0.684 & 6.76  & 0.038 & 0.163 & 4.30 \\
    \bottomrule
    \toprule
    \multicolumn{7}{c}{Computing {$\boldsymbol{e}$}} \\
    \midrule
    GPU   & \multicolumn{3}{c}{RTX 2080 Ti} & \multicolumn{3}{c}{RTX 3090} \\
    input  & int8(ms) & fp32(ms) & speedup & int8(ms) & fp32(ms) & speedup \\
    224   & 10.985 & 96.088 & 8.75  & 4.731 & 17.765 & 3.75 \\
    112   & 3.091 & 24.509 & 7.93  & 1.269 & 4.337 & 3.42 \\
    56    & 1.191 & 6.804 & 5.71  & 0.453 & 1.448 & 3.20 \\
    28    & 0.343 & 2.284 & 6.65  & 0.185 & 0.362 & 1.96 \\
    14    & 0.228 & 0.992 & 4.35  & 0.117 & 0.150 & 1.29 \\
    \bottomrule
    \toprule
    \multicolumn{7}{c}{Computing {$\boldsymbol{g}$}} \\
    \midrule
    GPU   & \multicolumn{3}{c}{RTX 2080 Ti} & \multicolumn{3}{c}{RTX 3090} \\
    input  & int8(ms) & fp32(ms) & speedup & int8(ms) & fp32(ms) & speedup \\
    224   & 365.696 & 535.745 & 1.46  & 122.301 & 462.291 & 3.78 \\
    112   & 109.723 & 132.783 & 1.21  & 27.327 & 130.079 & 4.76 \\
    56    & 27.611 & 43.139 & 1.56  & 6.911 & 32.083 & 4.64 \\
    28    & 5.061 & 10.695 & 2.11  & 1.074 & 0.458 & 0.43 \\
    14    & 1.466 & 3.632 & 2.48  & 0.272 & 0.178 & 0.65 \\
    \bottomrule
    \end{tabular}%
  \label{tab:aeg-speedup}%
\end{table}%

\subsection{FPGA Acceleration}
To evaluate the hardware benefits of using \type{int8} operations in NITI, an FPGA accelerator prototype was implemented using an Xilinx Alveo U280 FPGA card with the Xilinx Vitis HLS tools.  

\begin{figure}[htb]
    \begin{center}
       \includegraphics[width=1.0\linewidth]{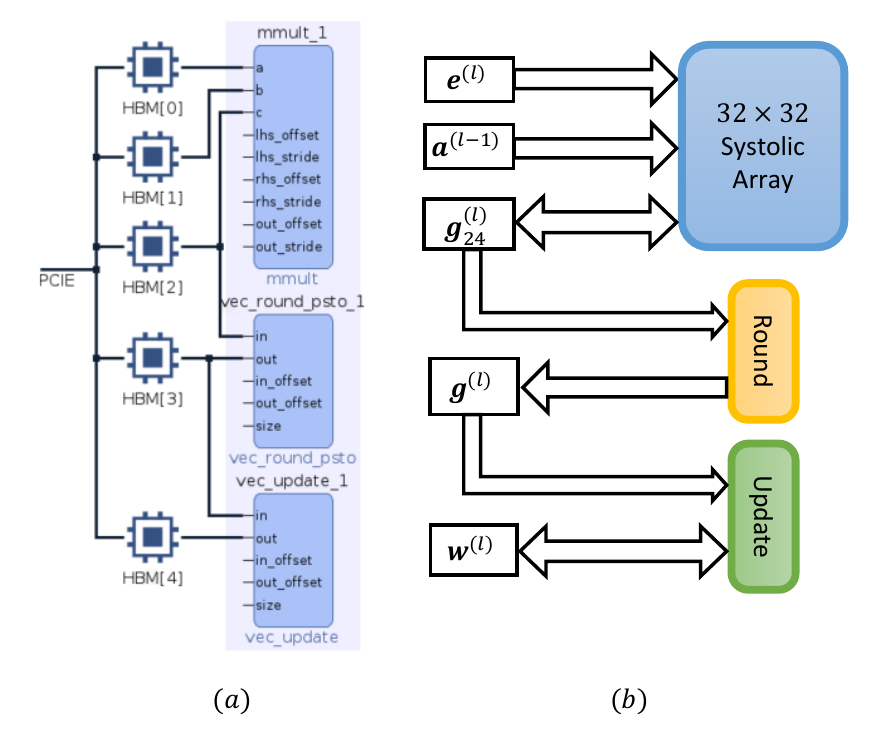}
    \end{center}
       \caption{(a)preliminary FPGA prototype of training a fully connected layer 
       \\(b)the data flow of computing $\boldsymbol{g}$ and updating $\boldsymbol{w}$ on the prototype}
    \label{fig:hls-hw}
\end{figure}

\figref{fig:hls-hw}(a) shows the top-level block diagram of the design.  The current design implements the 3 main operations of NITI, namely matrix multiplication, rounding, and weight update, needed to perform back propagation of a layer.
The layer weights, input, gradients and intermediate matrix multiplication accumulations during the training process are stored in the on-chip High Bandwidth Memory (HBM).

\figref{fig:hls-hw}(b) shows the data flow of computing $\boldsymbol{g}$ and updating $\boldsymbol{w}$ on the hardware. 
The data flow of computing $\boldsymbol{a}$ and $\boldsymbol{e}$ are similar and are omitted for brevity.
To effectively evaluate the hardware efficiency of each individual module when different arithmetic operations are employed, data are passed between different modules through the HBM.

Matrix multiplications form the core computational kernel of NITI.  Here, a $32 \times 32$ output-stationary systolic array was implemented in the hardware, which performs the operations for $1$ tile of the overall tiled matrix multiplications that compute $\boldsymbol{a}$, $\boldsymbol{e}$ and $\boldsymbol{g}$.
Synthesis results with different data types are shown in \tabref{table:mm-hls-synth}.
Considering the core systolic array for matrix multiplication shown in the first $2$ columns of the \tabref{table:mm-hls-synth}, when compared to the equivalent \type{fp32} implementation, the \type{int8} implementation consumes $5\times$ fewer DSP blocks and is able to operate at \SI{46}{\percent} higher clock speed.
It also incurs a kernel computation latency that is $12.1\times$ lower.

\begin{table}[tb]
  \centering
  \caption{Synthesis comparison of \type{fp32} and \type{int8}}
  \label{table:mm-hls-synth}
    \begin{tabular}{ccccc}
    \toprule
    Module      & \multicolumn{2}{c}{Systolic Array} & \multicolumn{2}{c}{Update} \\
    Precision & fp32  & int8  & \multicolumn{1}{c}{fp32} & \multicolumn{1}{c}{int8} \\
    \midrule
    Latency & 448 cycles & 37 cycles & \multicolumn{1}{c}{ 10 cycles} & \multicolumn{1}{c}{ 3 cycles} \\
    Frequency & 411.0 MHz & 601.3 MHz & \multicolumn{1}{c}{427.0MHz} & \multicolumn{1}{c}{520.3 MHz} \\
    LUT   &   299,091 & 18,531 & 13,280  & 7,424      \\
    FF    &   609,469 & 24,656 & 16,032  & 2,592   \\
    DSP   & 5120 (56\%) & 1024 (11\%) & 0     & 0 \\
    BRAM  & 6  &  6 & 0      & 0  \\
    \bottomrule
    \end{tabular}%
  \label{tab:addlabel}%
\end{table}%

\begin{table}[tb]
    \begin{center} 
    \caption{Synthesis Comparison of different rounding schemes } 
    \label{tab:round-hls-synth}
        \begin{tabular}{cccc} 
            \toprule
            Rounding  & Nearest & Stochasic  & Pseudo \\
            \midrule
            Latency & 3 cycles & 3 cycles & 3 cycles\\
            Clock & 520.3MHz & 520.3MHz & 520.3MHz \\
            LUT   & 3,328  & 4,928 & 3,680  \\
            FF    & 608  & 1,728 & 640  \\
            \bottomrule
            \multicolumn{4}{l}{No DSP or BRAM Usage}
        \end{tabular} 
    \end{center}
\end{table}

As mentioned in section \ref{sec:rounding}, the use of a proper rounding scheme is crucial to maintain training accuracy.
Here we evaluated the hardware cost of implementing different rounding schemes.
\tabref{tab:round-hls-synth} compares the hardware cost of running $32$ rounding operators in parallel using different schemes.
Although our proposed pseudo stochastic rounding schemes consumes more LUT and FF resources than the simple round to nearest scheme, it consumes \SI{63.0}{\percent} fewer FFs and \SI{25.3}{\percent} fewer LUT than the commonly used stochastic rounding scheme.

Finally, although simple in design, the weight update module similarly illustrates the benefit of \type{int8} operations.  As shown in \tabref{table:mm-hls-synth}, updates of weight is 3-cycle operation in \type{int8}.  On the other hand, due to the large trip time for the floating point operator, the \type{fp32} update modules requires $1.79\times$ more LUT and $2.86\times$ FF, while achieving a lower clock speed and $3.3\times$ more cycles latency.

\section{Conclusions}\label{sec:discuss}
In this work, we demonstrated the benefits of training deep neural networks using integer arithmetic exclusively in NITI.  The innovations of this work include: our update scheme that utilizes low precision integer operations for all intermediate values; a new stochastic rounding scheme that uses discarded bits as a random number source, obviating the need for an additional random number generator; and an efficient integer-only cross-entropy loss backpropagation scheme that is suitable for use in other integer DNN training frameworks.  By using $8$-bit integer arithmetic operations exclusively, we demonstrated that DNN training can be accelerated with existing low bitwidth integer operators in modern GPU originally designed for inference acceleration.  Preliminary study in the design of an FPGA based training accelerator also shows significant hardware advantages of \type{int8} implementations over \type{fp32} in terms of resource consumptions, memory storage requirements, as well as computation latency.  Looking into the future, our work lays the foundation for an integer-only accelerator which could greatly reduce cost, chip area and energy consumption required for DNN training.

\ifCLASSOPTIONcompsoc
  \section*{Acknowledgments}
\else
  \section*{Acknowledgment}
\fi
This work was supported in part by the Croucher Foundation Croucher Innovation Award 2013 and by ACCESS -- AI Chip Center for Emerging Smart Systems, sponsored by Innovation and Technology Fund (ITF), Hong Kong SAR.

\ifCLASSOPTIONcaptionsoff
  \newpage
\fi



%

\bibliographystyle{IEEEtran}
\bibliography{IEEEabrv, egbib, niti_hso}

\begin{IEEEbiography}
    [{\includegraphics[width=1in,height=1.25in,clip,keepaspectratio]{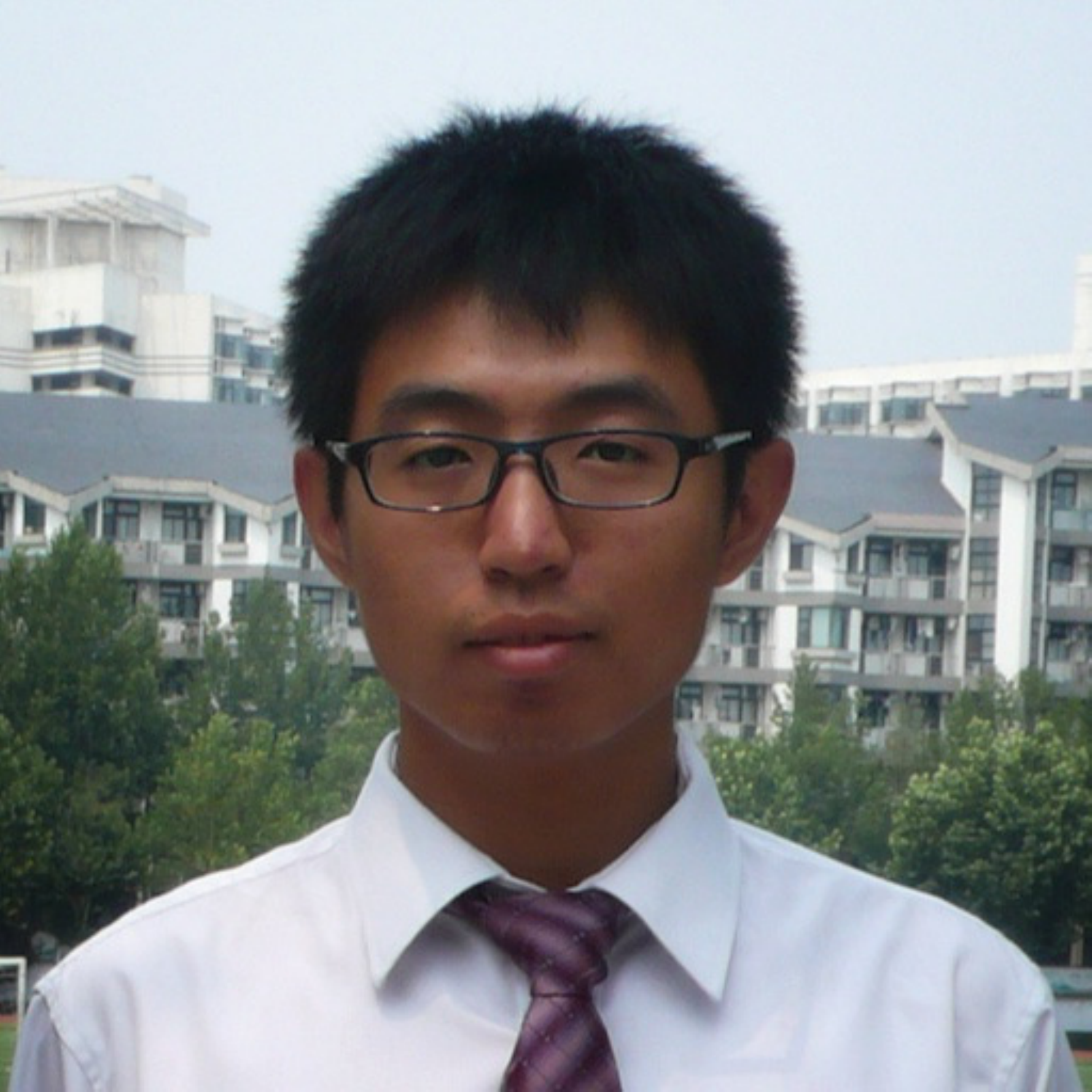}}]{Maolin Wang}
    received the B.Eng. degree in Electrical Engineering from Tsinghua University in 2015. 
    He then obtained his Ph.D. degree in the Department of Electrical and Electronic Engineering at The University of Hong Kong in 2020. 
    His research interests include efficient hardware architectures and algorithms for the training and inference of deep neural networks.
\end{IEEEbiography}
\begin{IEEEbiography}
[{\includegraphics[width=1in,height=1.25in,clip,keepaspectratio]{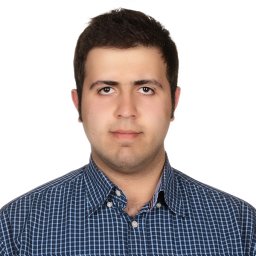}}]{Seyedramin Rasoulinezhad}
 is a PhD student researching on Computer architectures at The University of Sydney, Australia, under the supervision of Prof. Philip Leong and Dr David Boland. He obtained the B.S. degree in Electrical Engineering, Digital Systems at Sharif University of Technology, Iran. His research focus is on new FPGA architectures for the implementation of a variety of applications while considering machine learning algorithms as high demand future applications.
\end{IEEEbiography}

\begin{IEEEbiography}
[{\includegraphics[width=1in,height=1.25in,clip,keepaspectratio]{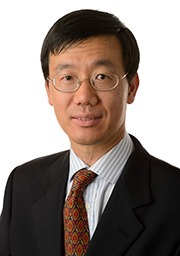}}]{Philip Leong}
received the B.Sc., B.E. and Ph.D. degrees from the University of Sydney. In 1993 he was a consultant to ST Microelectronics in Milan, Italy working on advanced flash memory-based integrated circuit design. From 1997-2009 he was with the Chinese University of Hong Kong. He is currently Professor of Computer Systems in the School of Electrical and Information Engineering at the University of Sydney, Visiting Professor at Imperial College and the Chief Technology Advisor to Cluster Technology. He was the co-founder and program co-chair of the International Conference on Field Programmable Technology (FPT); program co-chair of the International Conference on Field Programmable Logic and Applications (FPL) and is the Senior Associate Editor for the ACM Transactions on Reconfigurable Technology and Systems. The author of more than 150 technical papers and 4 patents, Dr. Leong was the recipient of the 2005 FPT conference Best Paper as well as the 2007 and 2008 FPL conference Stamatis Vassiliadis Outstanding Paper awards.
\end{IEEEbiography}

\begin{IEEEbiography}
    [{\includegraphics[width=1in,height=1.25in,clip,keepaspectratio]{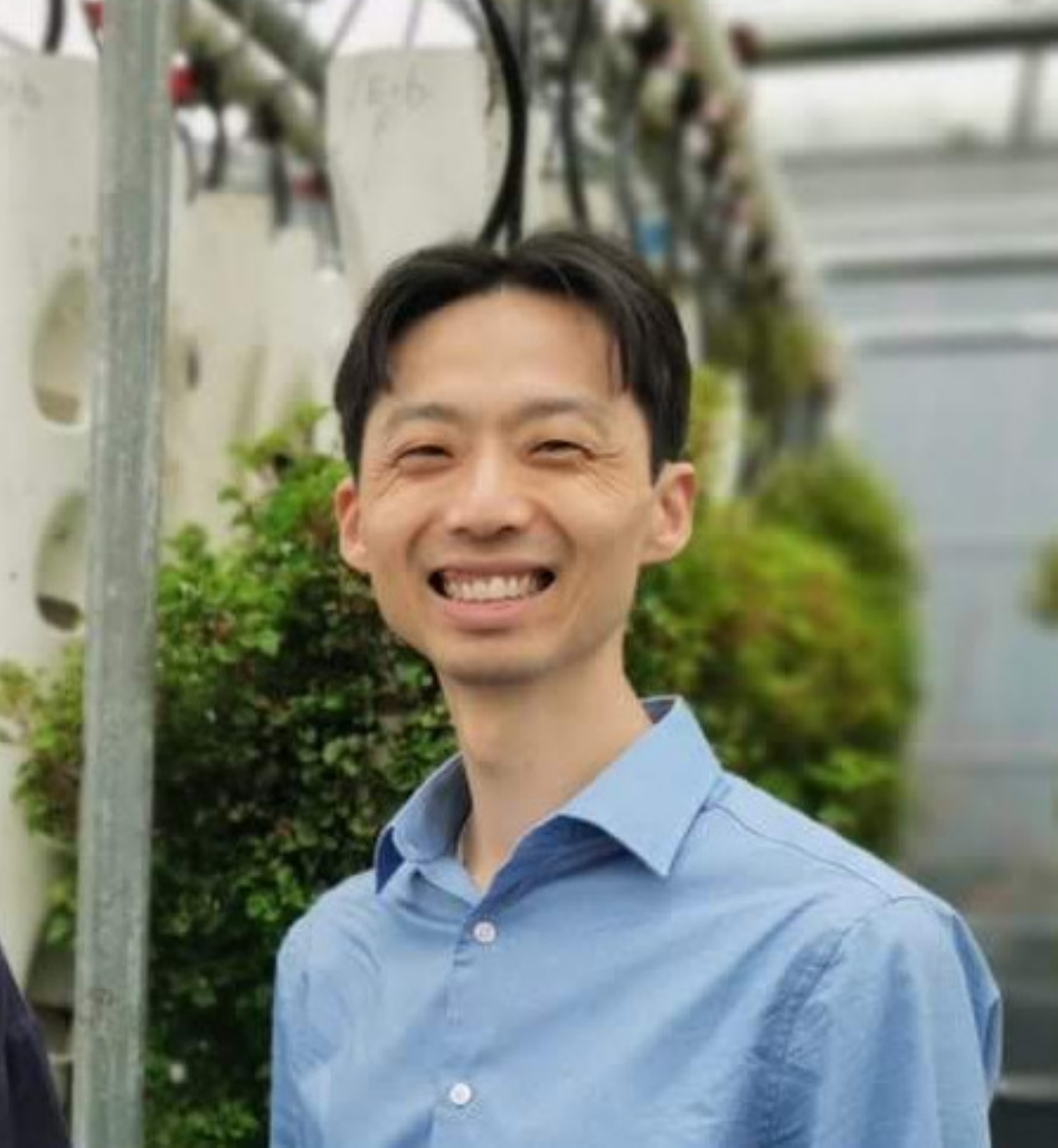}}]{Hayden K.-H. So}
 (S’03–M’07–SM’15) received the B.S., M.S., and Ph.D. degrees in electrical engineering and computer sciencesm from the University of California, Berkeley, CA, USA, in 1998, 2000, and 2007, respectively. He is currently an Associate Professor and Co-director of the computer engineering program at the Department of Electrical and Electronic Engineering, and the Co-director of the Joint Lab on Future Cities at the University of Hong Kong.  His research focuses on highly-efficient reconfigurable computing systems and their applications.  He was the recipient of the Test-of-time Award CODES+ISSS, in 2021, the Croucher Innovation Award, in 2013, the IEEE-HKN C. Holmes MacDonald Outstanding Teaching Award, in 2021, the University Outstanding Teaching Award (Team), in 2012, and the Faculty Best Teacher Award, in 2011. He has served as the technical program chair for various international conferences, including the International Symposium on Applied Reconfigurable Computing (ARC), the International Conference on Field-Programmable Technology (FPT), the International Symposium on Highly-Efficient Accelerators and Reconfigurable Technologies (HEART), and the IEEE International Conference on Application-Specific Systems, Architectures, and Processors (ASAP).
\end{IEEEbiography}

\end{document}